%% file: 0_main.tex
\documentclass{article} 
\usepackage{iclr2026_conference,times}

\input{math_commands.tex}

\usepackage{hyperref}
\usepackage{url}

\usepackage{wrapfig}
\usepackage{multirow}
\usepackage{booktabs}
\usepackage{grffile}  
\usepackage{subcaption} 
\usepackage{graphicx} 

\title{Disentanglement of Variations with Multimodal Generative Modeling}

\author{Yijie Zhang,\ \ Yiyang Shen,\ \ Weiran Wang \\
Department of Computer Science\\
University of Iowa\\
Iowa City, IA 52242, USA \\
\texttt{\{yijie-zhang,yiyang-shen,weiran-wang\}@uiowa.edu} \\
}

\iclrfinalcopy 
\begin{document}

\maketitle

\begin{abstract}
Multimodal data are prevalent across various domains, and learning robust representations of such data is paramount to enhancing generation quality and downstream task performance.
To handle heterogeneity and interconnections among different modalities, recent multimodal generative models extract shared and private (modality-specific) information with two separate variables. 
Despite attempts to enforce disentanglement between these two variables, these methods struggle with challenging datasets where the likelihood model is insufficient.
In this paper, we propose Information-disentangled Multimodal VAE (IDMVAE) to explicitly address this issue, with rigorous mutual information-based regularizations, including cross-view mutual information maximization for extracting shared variables, and a cycle-consistency style loss for redundancy removal using generative augmentations.
We further introduce diffusion models to improve the capacity of latent priors.
These newly proposed components are complementary to each other. Compared to existing approaches, IDMVAE shows a clean separation between shared and private information, demonstrating superior generation quality and semantic coherence on challenging datasets.
\end{abstract}

\input{1_intro}
\input{2_method}

\input{3_related}

\input{4_experiments}
\input{5_conclusion}
\newpage
\bibliography{multiview}
\bibliographystyle{iclr2026_conference}

\newpage
\appendix

\input{6_appendix}

\end{document}

%% file: math_commands.tex
\usepackage{amsmath,amsfonts,bm}

\def\eqref#1{equation~\ref{#1}}

\def\1{\bm{1}}

\def\rvw{{\mathbf{w}}}
\def\rvx{{\mathbf{x}}}

\def\rvz{{\mathbf{z}}}

\def\rmX{{\mathbf{X}}}

\DeclareMathAlphabet{\mathsfit}{\encodingdefault}{\sfdefault}{m}{sl}
\SetMathAlphabet{\mathsfit}{bold}{\encodingdefault}{\sfdefault}{bx}{n}

\def\gL{{\mathcal{L}}}

\newcommand{\E}{\mathbb{E}}

\newcommand{\KL}{D_{\mathrm{KL}}}

\newcommand{\norm}[1]{\left\lVert#1\right\rVert}

%% file: 1_intro.tex
\section{Introduction}
\label{sec:intro}
\vspace*{-1ex}

Most real-world data are inherently multimodal or multi-view\footnote{We use ``modality'' and ``view'' interchangeably as they both appear in the literature.}. Videos contain both visual scenes and sounds~\citep{zhao2018sound,owens2018audiovisual,chen2020vggsound,gong2023contrastive,kim2024equiav}; robots can see and feel via sensors~\citep{lee2019making}; images are often accompanied by captions~\citep{radford2021learning,jia2021scaling}; and heterogeneous human, animal, and environmental data are collected for health improvements~\citep{onehealth2022}. 
In addition to these naturally occurring data, synthetic multi-view data constructed from semantically similar input components or via augmentation are also widely used to learn useful representations for downstream tasks~\citep{velickovic2018deep,chen2020simple,caron2020unsupervised,tian2020contrastive,bardes2022vicreg}. Despite the abundance of such data, leveraging them is nontrivial even with naturally aligned modalities due to their diversity and complex correlations. Therefore, a core challenge is to integrate information across views to learn universal, transferrable representations. 

Variational autoencoders (VAEs,~\citealp{KingmaWellin14b}) and their multmodal extensions have emerged as a powerful paradigm to tackle this problem~\citep{wang2016vcca,suzuki2016joint}. They can extract useful shared information in data with missing modalities~\citep{wu2018multimodal} and noise~\citep{shi2021relating}. While early works have assumed that a single latent space can capture all relevant information and data variations~\citep{shi2019variational,sutter2021generalized}, recent approaches have recognized the existence of both shared and modality-specific (private) information in real-world datasets~\citep{daunhawer2022limitation,lee2021privateshared,palumbo2023mmvaeplus,palumbo2024deep}. 
However, modeling shared and private components naturally exposes a challenge: 
\vspace{-0.5em}
\begin{quote}
\textbf{\textit{How can we achieve maximal disentanglement between shared and private variables so that learned representations are complete and non-redundant?}}
\end{quote}
\vspace{-0.5em}
Without a clean separation, shared information leaks into private encodings and vice versa, causing weak coherence across modalities, wasted model capacity, and inadequate generative quality.

\paragraph{Our Contributions.} We propose Information-disentangled Multimodal VAE (IDMVAE), a novel framework for unsupervised multimodal representation learning hailing from theoretical stringency and practical performance.
(1) Different from prior works which use capacity-based~\citep{wang2016vcca} or shortcut-preventing~\citep{palumbo2023mmvaeplus} heuristics, we employ mutual information (MI)-based regularizations to ensure disentanglement. In particular, we use cross-view MI to extract common factors that likelihood models fail to fully capture, thereby enhancing cross-modal coherence. Additionally, a cycle-consistency-style loss removes redundancy between shared and private latents using samples generated by the model itself, eliminating the need for domain-specific augmentations.
(2) To overcome limitations of simple Gaussian priors, we leverage diffusion-based priors~\citep{dickstein15deep,ho2020denoising,song2021scorebased} that capture the richness of multimodal latent spaces, leading to greater representational capacities.
(3) Across multiple complex datasets spanning image, text and multi-omics data, IDMVAE performs consistently better than state-of-the-art methods in terms of cross-modal generation and coherence, showing a synergy between MI-based objectives and diffusion priors, which leads to improved performance.

%% file: 2_method.tex
\section{Method} 
\label{sec:method}
\vspace*{-2ex}

\begin{wrapfigure}{l}{0.45\linewidth}
\vspace*{-2ex}
\centering
\includegraphics[width=\linewidth]{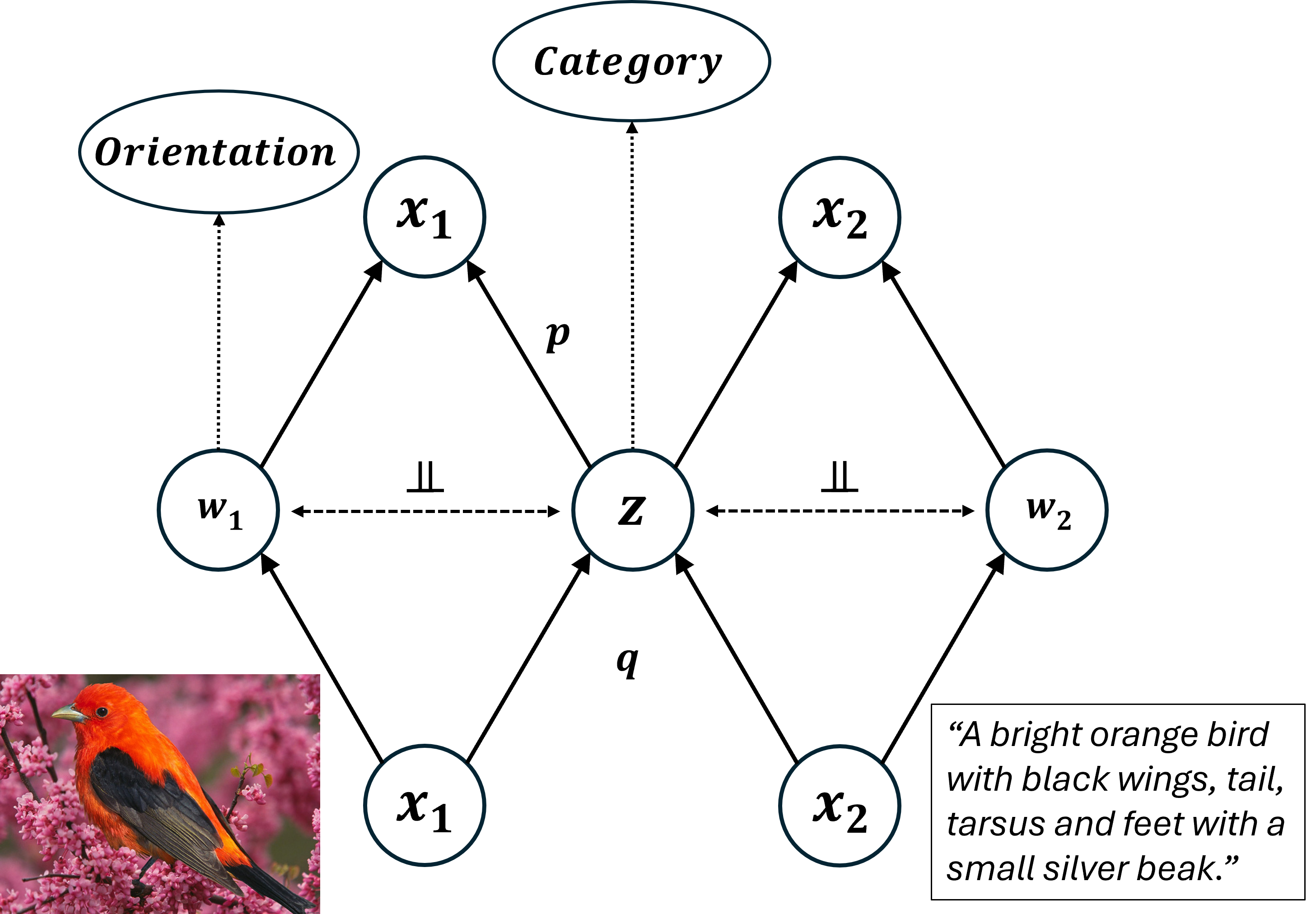}
\vspace{-4ex}
\caption{
Our approach with two modalities.}
\label{fig:schematic}
\vspace{-2ex}
\end{wrapfigure}

\vspace*{-0.5ex}
Given a set of $M$ modalities $\rmX = \{\rvx_1, \dots, \rvx_M\}$, we would like to learn a factorized latent space for each modality $m$, separating information shared across modalities, captured by latent variable $\mathbf{z}$, from modality-specific private information, captured by $\rvw_m$.
As illustrated in Figure~\ref{fig:schematic}, we assume that, $\forall m$, $\rvx_m$ is generated by $\rvz_m$ and $\rvw_m$ jointly with $p(\rvx_m|\rvz_m,\rvw_m)$, and the latent variables have independent priors $p(\rvz,\{\rvw_m\}_{m=1}^M) = p(\rvz)\prod_{m=1}^M p(\rvw_m)$.
We perform variational inference and parameterize the approximate posteriors with a factorized form: $q(\rvz,\rvw_m |\rvx_m)=q(\rvz|\rvx_m) \cdot q(\rvw_m|\rvx_m)$. Learned $q(\rvz|\rvx_m)$ and $q(\rvw_m|\rvx_m)$ provide representations of original inputs which can be used in 
downstream tasks. 
Our method consists of three major components.

\subsection{Likelihood modeling with Multimodal VAE}
\vspace*{-1ex}

We begin with the Evidence Lower Bound (ELBO) of the MMVAE+~\citep{palumbo2023mmvaeplus} model as our foundation. Let $q (\mathbf{z}|\rvx_m)$ be the posterior of shared variable derived from modality $m$ (with distribution parameters modeled by an encoder), and $q (\rvw_m|\rvx_m)$ be the posterior of private variable from modality $m$ (modeled by another encoder). One can 
define a global posterior of $\rvz$ by aggregating information from view-specific posteriors, e.g., using the mixture-of-experts (MoE) scheme $q (\mathbf{z}|\rmX) = \frac{1}{M}\sum_{m=1}^M q (\mathbf{z}|\rvx_m)$. Together with generative distributions $p (\rvx_m|\rvz, \rvw_m)$ modeled by decoders of each view, an 
ELBO for $\rmX$ can be derived with 
variational inference~\citep{KingmaWellin14b},  which involves reconstructing $\rvx_n$ using samples of $q (\rvw_n|\rvx_n)$ and $q (\mathbf{z}|\rvx_m)$, for $n,m=1,\dots,M$. 

Since the posterior $q (\rvw_n|\rvx_n)$ is dependent on $\rvx_n$ and potentially retains shared information, to keep decoder $p (\rvx_n|\rvz, \rvw_n)$ from taking the ``shortcut" to use leaked shared information, MMVAE+ uses sample of $\tilde{\rvw}_n$ from an auxiliary prior $r(\tilde{\rvw}_n)$ instead
for cross-view reconstruction, where $\rvz$ is derived from another view $m\neq n$; avoiding such shortcut enforces shared information to come from $q(\rvz|\rvx_m)$.
This design leads to the following minimizing objective:
\begin{align}
\mathcal{L}_{\text{MMVAE+}} = - \frac{1}{M} \sum_{m=1}^{M} \mathbb{E}_{\substack{\rvz \sim q(\rvz|\rvx_m) \\ \rvw_m \sim q(\rvw_m|\rvx_m) \\ \{\tilde{\rvw}_n \sim r(\tilde{\rvw}_n)\}_{n\neq m}}} 
\left[ \log \left( \frac{p (\rvx_m | \mathbf{z}, \rvw_m) p(\mathbf{z}) p(\rvw_m)}{q (\mathbf{z}|\rmX) q (\rvw_m|\rvx_m)} \prod_{n \neq m} p (\rvx_n|\mathbf{z}, \tilde{\rvw}_n) \right) \right].
\nonumber
\end{align}
Within the expectation, $p (\rvx_m | \mathbf{z}, \rvw_m)$ and $p (\rvx_n|\mathbf{z}, \tilde{\rvw}_n)$ are conditional likelihood of self- and cross-reconstructions, respectively. ~\citet{palumbo2023mmvaeplus} shows that $- \mathcal{L}_{\text{MMVAE+}}(\rvx_{1:M}) \le \log (\rvx_{1:M})$ remains a valid ELBO. Compared with previous multimodal VAE using MoE parameterization~\citep{shi2019variational}, MMVAE+ achieves superior performance for extracting shared information, although the result is somewhat sensitive to the relative capacity (dimensionality) of $\rvz$ and $\rvw$. 
In the rest of this section, we will show that regularizing the generative model with mutual information (MI) is more effective at achieving disentanglement.
We emphasize that MMVAE+ is one option for multimodal VAE, and our improvement below can also be applied to other models, e.g., PoE~\citep{wu2018multimodal} or MoPoE~\citep{sutter2021generalized}.


\subsection{Shared Variable Extraction with Cross-view MI Maximization}
\vspace*{-1ex}

While general (per-dimension) disentanglement of variations is theoretically challenging~\citep{locatello2019challenging}, the underlying structure of our setup that inputs of different modalities share a common cause facilitate (variable-level) disentanglement of shared versus private 
information. 

To extract the shared information, it is natural to enforce the shared representation of modality $m$, denoted $\rvz_m$ (with distribution $q(\rvz|\rvx_m$)) to have high mutual information (MI) with $\rvx_n$ for $n\neq m$.
Note this is partially pursued by $\mathcal{L}_{\text{MMVAE+}}$ through cross-view reconstruction. In light of the decomposition $I(\rvz_m, \rvw_n; \rvx_n) = I (\rvz_m; \rvx_n) + I (\rvw_n; \rvx_n | \rvz_m)$, we can maximize $I(\rvz_m, \rvw_n; \rvx_n)$ while minimizing $I (\rvw_n; \rvx_n | \rvz_m)$ to maximize $I (\rvz_m; \rvx_n)$. Focusing on the first term, 
since $I(\rvz_m, \rvw_n; \rvx_n) = H(\rvx_n) - H(\rvx_n|\rvz_m, \rvw_n)$ where the entropy $H(\rvx_n)$ is a constant,
minimizing the conditional entropy 
$H(\rvx_n|\rvz_m,\rvw_n) = \E_{\rvx_n,\rvz_m,\rvw_n} [- \log p(\rvx_n|\rvz_m,\rvw_n)]$
is equivalent to maximizing conditional likelihood. However, maximizing this upper bound does not ensure maximal $I(\rvz_m; \rvx_n)$ due to the gap $I(\rvw_n; \rvx_n|\rvz_m)$.
Therefore, likelihood maximization alone does not ensure disentanglement.

We thus take the alternative approach to maximize $I(\rvz_m, \rvz_n)$ which is a lower bound of $I(\rvz_m, \rvx_n)$:
$I(\rvz_m;\rvx_n)
    =I(\rvz_m; \rvz_n, \rvx_n) - I(\rvz_m; \rvz_n|\rvx_n)
    =I(\rvz_m; \rvz_n, \rvx_n) 
    =I(\rvz_m; \rvz_n)+I(\rvz_m;\rvx_n|\rvz_n)
    \geq I(\rvz_m;\rvz_n)$, 
where $I(\rvz_m;\rvz_n|\rvx_n)=0$ in the first step due to variability of $\rvz_n$ coming from $\rvx_n$ only~\citep{federici2020learning}.
In this work, we use the contrastive estimate of MI~\citep{oord2018representation}:
\begin{align} \label{eqn:contrastive-mi}
I(\rvz_m;\rvz_n) \approx 
Contrast(\rvz_m,\rvz_n) := \E_{\rvz_m,\rvz_n} \log \left[\frac{\phi(\rvz_m,\rvz_n)}{\phi(\rvz_m,\rvz_n) + \sum_{j=1}^k \phi(\rvz_m,\bar{\rvz}_n^j)}\right]
\end{align}
where $\phi(\rvz_m, \rvz_n) = \exp \left( \frac{\rvz_m^\top \rvz_n}{\norm{\rvz_m} \cdot \norm{\rvz_n}} \right)$ is the affinity function, and $\{\bar{\rvz}_n^j\}_{j=1}^k$ are $k$ negative examples randomly sampled from the minibatch not aligned with $\rvz_m$.
Since we have $M$ modalities, we compute the average of cross-modality MIs as our regularization for extracting shared information:
\begin{align*}
\mathcal{L}_{\text{CrossMI}} = - \frac{2}{M(M-1)} \sum\nolimits_{m < n} Contrast (\rvz_m, \rvz_n). 
\end{align*}

\subsection{Disentanglement with Generative Augmentation}
\label{sec:generative-augmentation}
\vspace*{-1ex}

While~\eqref{eqn:contrastive-mi} encourages $\rvz$ to capture shared information across views, it does not guarantee that the learned $\rvz_m$ contains no private information which should be modeled by $\rvw_m$. Similarly, even if $\rvz_m$ contains no private information and the self-reconstruction term in $\mathcal{L}_{\text{MMVAE+}}$ encourages $(\rvz_m,\rvw_m)$ to jointly capture all information about $\rvx_m$, the learned $\rvw_m$ can still retain shared information. Thus, we need additional regularization to remove redundancy between $\rvz_m$ and $\rvw_m$.

To motivate our method, consider the desired scenario where $\rvz_m$ and $\rvw_m$ are disentangled, so that they each can be varied independently to generate new samples of $\rvx_m$ using the decoder.
Let $\rvx_m$ and $\rvx_m^\prime$ be two input samples, and let $(\rvz_m, \rvw_m)$ be a pair of samples drawn from the posteriors $q(\rvz|\rvx_m)$ and $q(\rvw_m|\rvx_m)$, respectively, and similarly $(\rvz_m^\prime, \rvw_m^\prime)$ be a pair of samples drawn from conditional posteriors for $\rvx_m^\prime$.
With disentanglement and a good likelihood model, a sample $\rvx_m^+ \sim p(\rvx_m|\rvz_m,\rvw_m^\prime)$ would share the same $\rvz$ with $\rvx_m$. In turn, when we map $\rvx_m^+$ back to the latent space, $q(\rvz|\rvx_m^+)$ and $q(\rvz|\rvx_m)$ should be similar. 
Likewise, $q(\rvw_m|\rvx_m^+)$ and $q(\rvw_m|\rvx_m^\prime)$ should be similar.

More formally, assume that $\rvz_m$ is sufficient for $\rvx_n$, meaning that it captures all shared information, i.e., $I(\rvz_m; \rvx_n) = I(\rvx_m; \rvx_n)$ as encouraged by $\mathcal{L}_{\text{CrossMI}}$. Then in view of $I(\rvz_m; \rvx_n) = H (\rvz_m) - H (\rvz_m|\rvx_n)$, we would like to find the minimal $\rvz_m$ (with lowest $H(\rvz_m)$) by minimizing
\begin{align*}
    H (\rvz_m|\rvx_n)  = \E_{\rvz_m, \rvx_n} [- \log p (\rvz_m | \rvx_n)] \approx \E_{\rmX \sim p(\rmX), \rvz_m \sim q(\rvz|\rvx_m)} [- \log q (\rvz = \rvz_m | \rvx_n)] .
\end{align*}
Similar approaches have been used by~\citet[symmetric KL for minimizing $I(\rvx_m,\rvz_m|\rvx_n)$]{federici2020learning} and~\citet[inverse prediction]{tsai2019learning} for learning minimally sufficient shared variable. 
Essentially, for extracting shared information, $\rvx_m$ and $\rvx_n$ indeed constitute two views that are mutually redundant, satisfying $I(\rvx_m;\rvx_n|\rvz)=0$, so that the IB principle naturally applies. 

Note however, we do not have multiple natural views sharing $\rvw_m$ to carry out the above idea. This challenge motivates us to synthesize the view $\rvx_m^+ \sim p(\rvx_m|\rvz_m,\rvw_m^\prime)$, with $\rvw_m^\prime \sim q (\rvw|\rvx_m^\prime )$ from another sample $\rvx_m^\prime$, to reduce the redundancy of learned $\rvw_m$ by approximately minimizing $H (\rvw_m|\rvx_m^\prime)$ with the following loss
\begin{align*}
\E_{\rvx_m\sim p(\rvx_m), \rvx_m^\prime\sim p(\rvx_m), \rvz_m\sim q(\rvz|\rvx_m), \rvw_m^\prime\sim q(\rvw_m|\rvx_m^\prime), \rvx_m^+ \sim p(\rvx_m|\rvz_m,\rvw_m^\prime)}[-\log q(\rvw_m^\prime|\rvx_m^+)].
\end{align*}
Assuming that posteriors are parameterized Gaussians, this loss reduces to $\ell_2$ loss for matching means of posteriors.
In practice, we find it more stable to use a contrastive loss for matching, i.e.,
\begin{align*}
\gL_{\text{GenAug},\rvw_m} := - Contrast(\rvw_m^{\prime\prime}, \rvw_m^\prime)
\qquad\text{where}\quad 
\rvx_m^+ \sim p(\rvx_m|\rvz_m,\rvw_m^\prime),\;
\rvw_m^{\prime\prime} \sim q(\rvw_m | \rvx_m^+)
.
\end{align*}
We show results obtained with contrastive estimation in the main paper, and provide empirical comparisons of the two implementations in Appendix~\ref{sec:gen-aug-comparison}. 
We also define $\gL_{\text{GenAug},\rvz_m}$ similarly by switching the role of $\rvz_m$ and $\rvw_m$.
The total redundancy removal regularization is defined as
\begin{align*}
\gL_\text{GenAug}
= \frac{1}{2M}
\sum\nolimits_{m=1}^M \left( \gL_{\text{GenAug},\rvz_m} 
+ \gL_{\text{GenAug},\rvw_m} \right).
\end{align*}

Although we do not have the reconstruction target for $\rvx_m^+$, matching $q(\rvz|\rvx_m^+)$ with $q(\rvz|\rvx_m)$, and matching $q(\rvw_m|\rvx_m^+)$ with $q(\rvw_m|\rvx_m^\prime)$ implement a form of \emph{cycle-consistency}~\citep{zhu2017unpaired}, and provide learning signals for both the encoder and the decoder.
Previously,~\cite{bai2021contrastively} derived an ELBO of sequence data for disentangling static versus dynamic components, which involved mutual information terms based on data augmentation, similar to $\gL_\text{GenAug}$. However, their augmentation requires strong domain knowledge (e.g., shuffling the frame order does not alter the static component, and color change applied to all frames does not alter the dynamic component).
In contrast, our augmentations require no domain knowledge and are produced by the model itself.

\subsection{The Final IDMVAE Objective}

We define our objective of Information-Disentangled Multimodal VAE (IDMVAE) as 
\begin{equation}
\min \mathcal{L}_{\text{IDMVAE}} := \mathcal{L}_{\text{MMVAE+}} + \lambda_1 \mathcal{L}_{\text{CrossMI}} + \lambda_2 \mathcal{L}_{\text{GenAug}}
\label{eq:IDMVAE_final}
\end{equation}
where $\lambda_1$ and $\lambda_2$ are user parameters tuned on the validation set.

\paragraph{Diffusion Priors}
In most multimodal VAEs, the prior distributions are chosen to be simple and easy to sample from, e.g., Gaussian for continuous data. 

However, such unstructured priors may not be ideal for representation learning, whose purpose is to discover useful structure of data for supervised downstream tasks. As an example, a representation containing rich label information most likely have a clustering structure where data of different classes are separated far apart, and will not have a uni-modal distribution like Gaussian. 
We use diffusion models~\citep{dickstein15deep,ho2020denoising,song2021scorebased} to overcome this limitation by parameterizing $p(\rvz)$ as a denoising process started with pure noise. To naturally introduce diffusion models into our loss,
we decompose the KL divergence inside $\mathcal{L}_{\text{MMVAE+}}$ (which we minimize) as~\citep{vahdat2021score}:
\begin{gather} \label{eqn:diffusion-kl}
\KL ( q (\rvz|\rvx) || p(\rvz)) = E_{q (\rvz|\rvx)} \left[\log q (\rvz|\rvx) \right] + E_{q (\rvz|\rvx)} \left[ - \log p(\rvz) \right]. 
\end{gather}
The first term maximizes the entropy of the approximate posterior $q(\rvz|\rvx)$. The second term maximizes the likelihood of samples from $q(\rvz|\rvx)$ under $p(\rvz)$, which we model with diffusion models. We can treat $\rvz \sim q(\rvz|\rvx)$ as ``data'', and destroy its structure by gradually adding noise to it, resulting in pure noise after a number of steps. With repeated applications of a denoising network, diffusion models gradually reverse the noising process, and recover the original data from pure noise. Diffusion models have well-defined ELBO objectives which \emph{lower bound} $\log p(\rvz)$, and plugging them into~\eqref{eqn:diffusion-kl} yields \emph{valid upper bounds} of the KL divergence. 
Since the latent variables are of low dimensionality, we parameterize the diffusion backward process with a simple feedforward network.
In practice, we introduce additional loss weight for $E_{q (\rvz|\rvx)} \left[ - \log p(\rvz) \right]$, and model the mean of $q(\rvz|\rvx)$ with the DDPM parameterization~\citep{ho2020denoising}. 
We optimize over all modules (encoders, decoders, diffusion networks) jointly in an \emph{end-to-end} manner.
A recent work~\citep{palumbo2024deep} proposed a two-step approach which first learns the representations with MMVAE+, and then learns diffusion models in the input space, conditioned on 
VAE reconstructions. Note our use of diffusion model has a different motivation, and we jointly train it during representation learning.

%% file: 3_related.tex
\section{Related Work}
\label{sec:related_work}
\vspace*{-2ex}

\paragraph{Disentanglement in VAEs.} To achieve disentangled latent representations in VAEs, researchers have commonly used mutual information (MI) based regularization, and employed various metrics to assess the results~\citep{higgins2017betavae,kim2018disentangling,chen2018isolating,kumar2018variational}. However, it has been shown that, without supervision or inductive bias in the model, it is theoretically challenging to recover (per-dimension) disentanglement~\citep{locatello2019challenging}.

\vspace*{-1ex}
\paragraph{Contrastive and self-supervised learning (SSL).}
SSL is a paradigm that aims at learning useful representations from large amounts of unlabeled data, by creating \emph{artificial targets} that are generally correlated with downstream tasks. SSL is often applied to a single modality, with artificial views created based on the structures of data~\citep{oord2018representation,logeswaran2018efficient,hjelm2018learning,bachman2019learning,chen2020simple,caron2020unsupervised,tian2020contrastive,bardes2022vicreg,zbontar2021barlow},
as well as multimodal data~\citep{radford2021learning,jia2021scaling,elizalde2023clap}
and many methods are motivated by the classical infomax principle~\citep{linsker1988self} and they implement neural estimation of mutual information, with contrastive loss being the most popular variant.
Recent works have proposed theoretical interpretations of SSL and contrastive learning~\citep{wang2020understanding,Zimmermann_21a,hyvarinen2019nonlinear,tian2020what,tosh2021contrastive,chen2021provable,zhai2024understanding}, with the focus of providing guarantees for extracting the shared variable, without considering the private variables.

A few works took private variations into consideration. 
\cite{kugelgen2021self} proposed a generative model in which the latent space is divided into  ``content'' and ``style''; importantly, data augmentations were assumed to preserve  content  while altering dimensions within  style. 
~\cite{tsai2021selfsupervised} studied self-supervised learning from a multi-view perspective 
and with the \emph{multi-view redundancy} assumption~\citep{Chaudh_09a,tosh2021contrastive} that the private variable of each view contains little 
information for the downstream task, they focused on extracting the shared variable 
with combinations of several multi-view losses.
Realizing the limitation of this assumption,~\cite{liang2023factorized} studied the scenario where the private variables contain significant useful information, and proposed a contrastive learning algorithm for extracting it. Their algorithm required sophisticated data augmentation procedures designed for downstream task.
~\cite{lyu2022understanding} proposed a model for understanding SSL, assuming a data generation process similar to ours. 
They extracted shared variable with CCA loss, and 
private variable by MI minimization between shared and private variables. The generation quality of their model is suboptimal, due to it being fully deterministic.

\vspace*{-1ex}
\paragraph{Information bottleneck (IB) and mutual-information regularization.}

Another set of probabilistic models were motivated by the IB method~\citep{tishby1999information,tishby2015deep,achille2018emergence}.
\citet{Alemi_17a} proposed a variational IB method to extract $\rvz$ from $\rvx_1$ which has high MI with $\rvx_2$ (estimated with conditional likelihood), so that it captures the shared information, and at the same time has low MI with $\rvx_1$ so that it contains little nuisance factors/private information. 
\cite{federici2020learning} leveraged the multi-view redundancy assumption that all the information $\rvx_1$ contains about an unobserved label is also contained in $\rvx_2$, and showed that that if the learned representation $\rvz$ is sufficient, in the sense that $I(\rvx_1,\rvx_2|\rvz)=0$, then $\rvz$ has all the predictive power from $(\rvx_1,\rvx_2)$ for label. 
Remarkably, their objective did not involve any reconstruction paths, and the authors considered this to be an advantage, given that density modeling for high dimensional data is difficult. 
\citet{wang2025information} extended~\citet{federici2020learning} and proposed a two-step approach to first extract shared and then the private variables with guarantees, again without generative modeling.
We argue that, with the development of powerful generative models, likelihood modeling becomes feasible and provides the additional benefit of (controllable) generation. 

%% file: 4_experiments.tex
\section{Experiments}
\label{sec:experiments}
\vspace*{-2ex}
We compare our method, IDMVAE, and its variant with diffusion priors, against several baselines. 

\noindent \textbf{MMVAE} \citep{shi2019variational}: 
uses a 
MoE inference network to combine information from different modalities. It only models the shared variable $\rvz$ with ELBO.

\textbf{MoPoE-VAE} \citep{sutter2021generalized}: 
uses a mixture-of-products-of-experts inference network for $\rvz$.

\textbf{DMVAE} \citep{lee2021privateshared}: 
performs 
PoE inference for $\rvz$, and 
models $\rvw_m$ within ELBO.

\textbf{MMVAE+} \citep{palumbo2023mmvaeplus}: 
performs separation of shared versus private information with the help of auxiliary prior variables. 
It is a special case of IDMVAE (w.o. diffusion) with $\lambda_1=\lambda_2=0$.

\textbf{DisentangledSSL} \citep{wang2025information}: performs  extraction of shared variable (using the method of~\citealp{federici2020learning}) and private variable in two sequential steps. It is a state-of-the-art disentanglement method without likelihood modeling, but it can only be applied to two views currently.


%

\subsection{Results on PolyMNIST-Quadrant}
\label{sec:polymnist_results}
\vspace*{-1ex}

\begin{wrapfigure}{l}{0.44\linewidth}
\vspace*{-2.5ex}
\centering
\includegraphics[width=\linewidth]{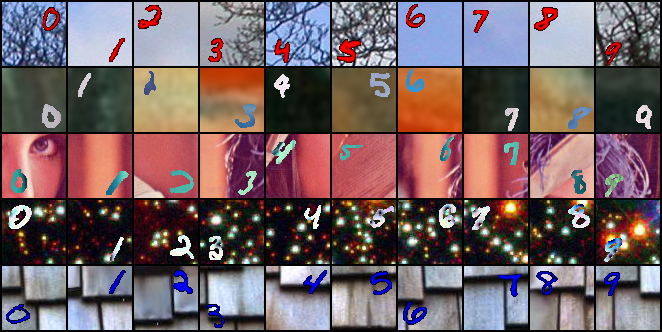}
\vspace*{-4ex}
\caption{PolyMNIST-Quadrant dataset. Digits (0-9) are placed in one of the four quadrants randomly.  Each column contains one multimodal sample.  Each modality has a different background scheme.
Digit label is shared across all modalities, while quadrant label is private to each modality.}
\label{fig:polymnist_quadrant_sample}
\vspace*{-3ex}
\end{wrapfigure}

PolyMNIST~\citep{sutter2021generalized} is a benchmark for multimodal representation learning, consisting of MNIST~\citep{Lecun_98a} digits overlaid on complex backgrounds. We make the dataset more challenging, by taking each MNIST digit and placing a 32x32
scaled version of it into one of four quadrants of a 64x64 canvas;
see Figure~\ref{fig:polymnist_quadrant_sample} for an illustration. 
This modification introduces the private latent variable which captures the quadrant position (with ground truth label) and background for each modality, allowing for nuanced evaluation of disentanglement and generation. 
Our training/validation/test sets contain 220,000/5,000/10,000 samples.
We use the deep residual network~\citep{he2015deepresiduallearningimage} architecture as the backbone of encoders and decoders for all methods.
{The dimensionality is set to 32 for $\rvz$ and 128 for $\rvw_m$.}


\begin{table}[t]
\centering
\caption{Latent classification on PolyMNIST-Quadrant. 
Accuracies are averaged over 5 modalities. 
}
\label{tab:polymnist_results_part1}
\vspace*{-1ex}
\begin{tabular}{l|cc|cc}
\hline
Model & $z \to Digit \uparrow$ & $z \to Quad \downarrow$ & $w \to Quad \uparrow$ & $w \to Digit \downarrow$ \\  %
\hline
MMVAE                       & 0.492         & 0.798         & ---   & ---       \\
MoPoE-VAE                   & 0.536         & 0.751         & ---   & ---       \\
DMVAE                       & 0.157         &\textbf{0.254} & 0.710 & 0.179     \\ 
MMVAE+                      & 0.382         & 0.355         & 0.999 & 0.341     \\ 
\hline
\textbf{IDMVAE (ours)}      &\textbf{0.983} & 0.271         & 0.999 & 0.162     \\  
\hspace{0.8em} 
    – $\gL_{\text{CrossMI}}$ 
    ($\lambda_1=0$)         & 0.111         & 0.267         & 0.999 & 0.356     \\ 
\hspace{0.8em} 
    – $\gL_{\text{GenAug}}$ 
    ($\lambda_2=0$)         & 0.977         & 0.277         & 0.999 & 0.202     \\  
+ Diffusion prior           & 0.982         & 0.267         & 0.999 & \textbf{0.143}     \\  
\hline
\end{tabular}
\end{table}

\vspace*{-1ex}
\paragraph{Latent Classification.} 
For evaluation, we perform linear classification on the \emph{samples} of posterior distributions (samples reflect both mean and variance of posteriors). Multi-class logistic regression models are trained on the posterior samples of training set and applied to posterior samples of the test set.
We perform two types of classifications: (1) predicting shared label from the shared variable ($\rvz$) and private label from the private variable ($\rvw_m$), where high accuracy is better, indicating the desired variation is captured; and (2) cross-classification, where we predict shared label from $\rvw_m$ and predict quadrant label from $\rvz$. Ideally, with successful disentanglement, cross-classification accuracies should approach the performance of a random classifier (e.g., 10\% for predicting digits from $\rvw_m$, 25\% for predicting quadrants from $\rvz$).
We present results of different methods in Table~\ref{tab:polymnist_results_part1}, as well as performance of our method when either $\gL_{\text{CrossMI}}$ or $\gL_{\text{GenAug}}$ is removed from our loss.
Clearly, our method achieves superior performance. $\gL_{\text{CrossMI}}$ is critical for extracting the shared variable, and this is because the digits occupy a small number of pixels and pure likelihood modeling may ignore them. 
$\gL_{\text{GenAug}}$ helps remove redundant information, so that cross-classification accuracy is reduced.
Adding diffusion in latent space (last row of table) leads to small gain.


\begin{table}[t]
\centering
\caption{Generative coherence, averaged over 5 views, on PolyMNIST-Quadrant. We use subscript $q$ to indicate samples from posteriors and subscript $p$ to indicate samples from priors.
For generated images, digit label is determined by $\rvz_{s,q}$ or otherwise random (with target accuracy 10\%), quadrant label is determined by $\rvw_{s,q}$ or otherwise random (with target accuracy 25\%).  
}
\label{tab:polymnist_results_part2}
\vspace*{-1ex}
\begin{tabular}{@{}l||@{\hspace{0\linewidth}}c@{\hspace{0\linewidth}}c@{\hspace{0\linewidth}}|@{\hspace{0\linewidth}}c@{\hspace{0\linewidth}}c@{\hspace{0\linewidth}}||@{\hspace{0\linewidth}}c@{\hspace{0\linewidth}}c@{\hspace{0\linewidth}}||@{\hspace{0\linewidth}}c@{}}
\hline
\multirow{3}{*}{Model} & \multicolumn{4}{c@{\hspace{0.0\linewidth}}||}{Self Gen ($s=t$)} & \multicolumn{2}{c@{\hspace{0.0\linewidth}}||}{Cross Gen  ($s \neq t$)} & Uncond. \\
\cline{2-8}
& \multicolumn{2}{c|}{$Gen(\rvz_{s,q}, \rvw_{t,p})$} & \multicolumn{2}{c||}{$Gen(\rvz_{t,p}, \rvw_{s,q})$} & \multicolumn{2}{c||}{$Gen(\rvz_{s,q}, \rvw_{t,p})$} & $Gen(\rvz_{p}, \rvw_{p})$\\
\cline{2-8}
 & $Digit \uparrow$ & $ Quad \downarrow$ & $ Digit \downarrow$ & $ Quad \uparrow$ & $ Digit \uparrow$ & $ Quad \downarrow$ & $ Digit \uparrow$ \\
\hline
MMVAE       & 0.879 & 0.998 & ---   & ---   & 0.170 & 0.248 & 0.041 \\ %
MoPoE-VAE   & 0.861 & 0.999 & ---   & ---   & 0.173 & 0.250 & 0.029 \\
DMVAE       & 0.297 & 0.252 & 0.532 & 0.999 & 0.161 & 0.249 & 0.005 \\  %
MMVAE+      & 0.120 & 0.251 & 0.915 & 0.999 & 0.119 & 0.250 & 0.000 \\
\hline
\textbf{IDMVAE (ours)}                      & 0.898         & 0.249 & 0.162         & 0.999 & 0.881         & 0.250  & 0.070 \\
\hspace{0.8em} 
    – $\gL_{\text{CrossMI}}$ ($\lambda_1=0$)& 0.101         & 0.252 & 0.926         & 0.999 & 0.100         & 0.250  & 0.000 \\
\hspace{0.8em} 
    – $\gL_{\text{GenAug}}$ ($\lambda_2=0$) & 0.670         & 0.250 & 0.370         & 0.999 & 0.671         & 0.250  & 0.008 \\
+ Diffusion prior                           &\textbf{0.942} & 0.251 &\textbf{0.106} & 0.999 &\textbf{0.887} & 0.251  & \bf 0.664 \\
\hline
\end{tabular}
\end{table}

\begin{figure}[t]
    \centering
    \begin{tabular}{@{}c@{\hspace{0.02\linewidth}}c@{\hspace{0.02\linewidth}}c@{\hspace{0.02\linewidth}}c@{\hspace{0.02\linewidth}}c@{}}
        \begin{minipage}[t]{0.18\linewidth}
            \centering
            \includegraphics[width=\linewidth]{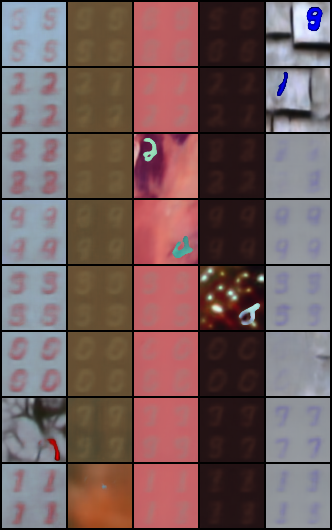}
            \vspace{1mm}
             MMVAE 
        \end{minipage} &
        \begin{minipage}[t]{0.18\linewidth}
            \centering
            \includegraphics[width=\linewidth]{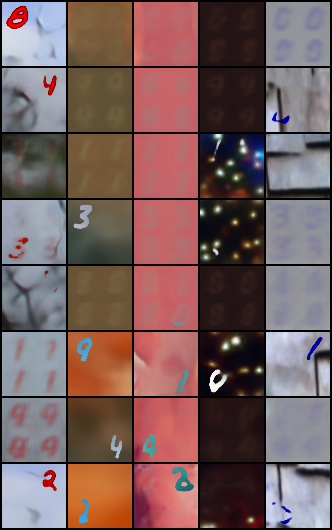}
            \vspace{1mm}
             MoPoE-VAE 
        \end{minipage} &
        \begin{minipage}[t]{0.18\linewidth}
            \centering
            \includegraphics[width=\linewidth]{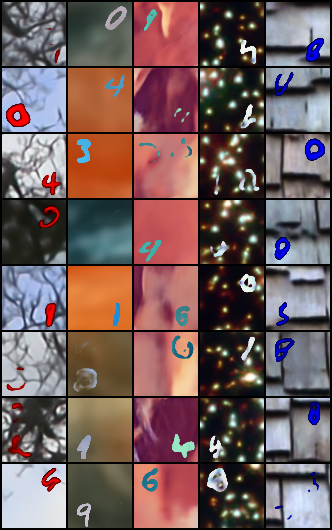}
            \vspace{1mm}
             DMVAE 
        \end{minipage} &
        \begin{minipage}[t]{0.18\linewidth}
            \centering
            \includegraphics[width=\linewidth]{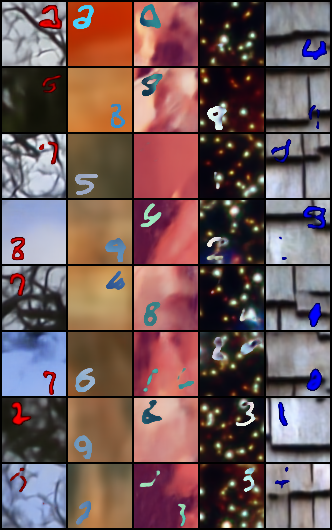}
            \vspace{1mm}
             MMVAE+ 
        \end{minipage} &
        \begin{minipage}[t]{0.18\linewidth}
            \centering
            \includegraphics[width=\linewidth]{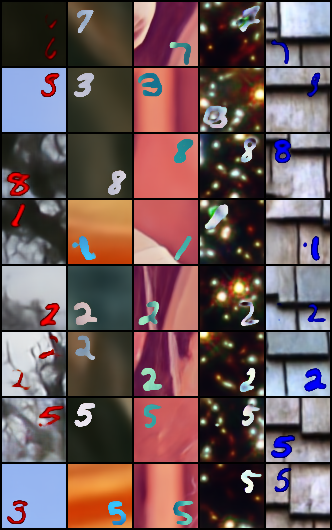}
            \vspace{1mm}
             IDMVAE + Diff.
        \end{minipage}
    \end{tabular}
    \vspace*{-1ex}
    \caption{Unconditional generations on PolyMNIST-Quadrant. Each row is a multimodal sample generated with a prior sample of $\rvz$, so images in the same row ideally have the same digit identity.
    }
    \label{fig:unconditional_comparison}
    \vspace*{-3ex}
\end{figure}

\vspace*{-1ex}
\paragraph{Conditional Coherence.} This metric evaluates the model's ability to generate consistent samples across modalities. We assess this for both self-reconstruction and cross-modal generation.
Formally, we combine either posterior $\rvz_{s,q} \sim q(\rvz|\rvx_s)$ or prior $\rvz_{s,p} \sim p(\rvz)$ (using diffusion prior if available) of a 
modality $s$, with
the posterior $\rvw_{t,q} \sim q(\rvw|\rvx_t)$ or the prior 
$\rvw_{t,p} \sim p(\rvw_t)$ (using diffusion prior if available) of modality $t$, and apply  $q(\rvx_t|\rvz,\rvw_t)$ to generate a new sample of modality $t$. This sample should have the same digit label as $\rvx_s$ if posterior of $\rvz$ is used, and random digit label if prior is used.
Similarly, the quadrant label can be determined based on whether posterior or prior is used for $\rvw_t$.
We then use ResNet classifiers 
trained on original images to predict corresponding labels of generated images, and the averaged accuracy across modalities is referred to as \emph{coherence}.
We provide conditional generative coherence in Table~\ref{tab:polymnist_results_part2} (left panel for self generation where $s=t$, and middle panel for cross generation $s\neq t$); see Appendix~\ref{sec:polymnist_quadrant_conditional_generation} for sample generations. The results are consistent with those of latent classification, and diffusion priors significantly boost coherence.

\vspace*{-3ex}
\paragraph{Unconditional Coherence.} 
This metric further assesses the consistency of the shared information in unconditionally generated samples. 
We first sample a shared latent code $\rvz_p \sim p(\rvz)$ (using diffusion prior when available). 
For each modality $m$, we then sample an independent private 
$\rvw_{m,p} \sim p(\rvw_m)$ and generate a sample $\hat{\rvx}_m$ from the combined latent code $(\rvz_p, \rvw_{m,p})$. The generated multimodal sample $\{\hat{x}_1, ..., \hat{x}_M\}$ are then passed to their respective digit classifiers (ResNet) trained on original training images, to predict the shared label. A sample set is considered coherent if \emph{all}  classifiers agree on the same shared label. We report the percentage of coherent sets as unconditional coherence, shown in Table~\ref{tab:polymnist_results_part2} (right panel). Most methods obtain close to zero unconditional coherence, indicating the difficulty of matching prior and posterior distributions for latent variables. 
However, with diffusion prior our method achieves significantly better coherence, thanks to its flexibility. 
We show generations in Figure~\ref{fig:unconditional_comparison}, and 2D visualizations of latent codes in Appendix~\ref{sec:polymnist-quadrant-latent-visualization}.

\vspace*{-4.5ex}
\begin{wrapfigure}{l}{0.28\linewidth}
\vspace*{-2ex}
\centering
\includegraphics[width=0.97\linewidth]{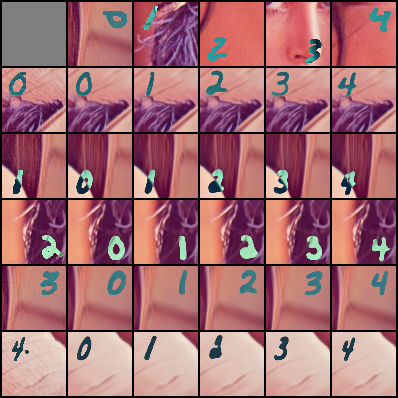}
\vspace{-2ex}
\caption{
Augmentations.}
\vspace{-4ex}
\label{fig:gen-aug-illustration}
\end{wrapfigure}

\paragraph{Generative augmentation.} Recall that in $\gL_{\text{GenAug}}$ we mix and match posteriors of $\rvz$ and $\rvw_m$ from different samples to generate new samples in modality $m$. We provide illustration of such samples from our trained model in Figure~\ref{fig:gen-aug-illustration}. The first row and first column contain images for which we extract posterior samples of $\rvz$ and $\rvw_m$ respectively. And the rest of the grid contain generate images using samples of $\rvw_m$ of the corresponding row and $\rvz$ of the corresponding column. We observe that images in each column share the same digit, while images in each row share share the same quadrant, as desired. Generated images are of high quality, showing that we can independently vary shared and private variables to obtain controllable generations.
%


\subsection{Results on CUB}
\label{sec:cub_results}
\vspace*{-1ex}

The CUB-200-2011 dataset~\citep{wah2011caltech, reed2016learning, shi2019variational, palumbo2023mmvaeplus, palumbo2024deep} 
is a widely used benchmark for fine-grained visual categorization, containing 64x64 RGB images of 200 bird species. Each image is paired with 10 textual descriptions. 
Following~\citet{palumbo2024deep}, we group 22 categories of species from the 200 bird species into 8 super-categories, yielding 1-of-8 class labels for these species. Data with category label is split into training/validation/test with 80\%/10\%/10\% portions. The rest 178 species are added to the training set for representation learning. The training/validation/test sets contain 115,240/1,280/1,360 samples, respectively. 
See more details on data generation in Appendix \ref{sec:dataset_cubcluster}.

For this dataset, the two modalities (image, text) share rich information about bird category, since the text describes the color of different parts of the bird. To evaluate the quality of private information, we note that the horizontal direction of the bird (with direction inferred from the original CUB attributes, see Appendix~\ref{sec:dataset_cubcluster} for details) can only be inferred from the image. Therefore, we consider the direction as private label for the image modality.

\begin{table}[t] 
\centering
\caption{Latent 
classification on CUB, using posterior means.  $\rvz_1$ and $\rvw_1$ refers to image latents. 
}
\label{tab:cub_results_part1}
\vspace*{-1ex}
\begin{tabular}{l|cc|cc}
\hline
Model & \textbf{$\rvz \to Cat. \uparrow$} & \textbf{$\rvz_1 \to Dir. \downarrow$} & \textbf{$\rvw_1 \to Dir. \uparrow$} & \textbf{$\rvw \to Cat. \downarrow$} \\
\hline
MMVAE                       & 0.685          & 0.820   & ---             & ---     \\  
MoPoE-VAE                   & 0.731          & 0.837   & ---             & ---     \\  
DMVAE                       & 0.418          & 0.771   & \bf{0.843}  & 0.400   \\  
MMVAE+                      & 0.725          & 0.692   & 0.612           & 0.323   \\  
DisentangledSSL             & 0.831     & 0.557   & 0.592           & \bf 0.179   \\  
\hline
IDMVAE \textbf{(ours)}      & 0.815     & \textbf{0.501}   & 0.720   & 0.200   \\  %
\hspace{0.8em} – 
    $\gL_{\text{CrossMI}}$ 
    ($\lambda_1=0$)         & 0.759   & 0.767   & 0.635   & 0.292   \\  
\hspace{0.8em} – 
    $\gL_{\text{GenAug}}$ 
    ($\lambda_2=0$)         & 0.810   & 0.493   & 0.698   & 0.230   \\  
+ Diffusion prior          & \textbf{0.840}   & 0.526   & 0.667   & 0.321   \\  
\hline
\end{tabular}
\end{table}

\begin{figure}[t]
    \centering
        \begin{tabular}{@{}
c@{\hspace{0.02\linewidth}}c@{\hspace{0.02\linewidth}}c@{\hspace{0.02\linewidth}}c
        @{\hspace{0.04\linewidth}}
c@{\hspace{0.02\linewidth}}c@{\hspace{0.02\linewidth}}c@{\hspace{0.02\linewidth}}c  
        @{}}

        \cmidrule(lr){1-4} \cmidrule(lr){5-8}
        & MMVAE+ & IDMVAE &  + Diffusion & 
        & MMVAE+ & IDMVAE & + Diffusion \\ 
        \cmidrule(lr){1-4} \cmidrule(lr){5-8}
        
        \raisebox{-3ex}{\includegraphics[width=0.09\linewidth]{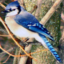}} & 
        \parbox[c]{0.08\linewidth}{\scriptsize this bird has blue blue blue blue and blue a blue beak .} &

        \parbox[c]{0.10\linewidth}{\scriptsize the bird has blue colored feathers and and has long . .} &

        \parbox[c]{0.12\linewidth}{\scriptsize this bird has a white head and a and throat and and and and blue black and black speckled .} &

        \raisebox{-3ex}{\includegraphics[width=0.09\linewidth]{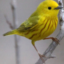}} & 
        \parbox[c]{0.08\linewidth}{\scriptsize this bird bird yellow and black and and and a short and . .} &

        \parbox[c]{0.10\linewidth}{\scriptsize the bird has yellow color feathers and thin two long eyering .} &

        \parbox[c]{0.12\linewidth}{\scriptsize the bird has a yellow belly , belly , belly and rectrice and a is black and black speckled .} \\[6ex]

        \raisebox{-3ex}{\includegraphics[width=0.09\linewidth]{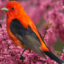}} & 
        \parbox[c]{0.08\linewidth}{\scriptsize this bird has red with with and and , and short beak .} &

        \parbox[c]{0.09\linewidth}{\scriptsize the bird has red breasted feathers and and a long . .} &

        \parbox[c]{0.11\linewidth}{\scriptsize this bird has a red head , a and throat and and and and is black and black beak .} &
        
        \raisebox{-3ex}{\includegraphics[width=0.09\linewidth]{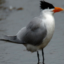}} & 
        \parbox[c]{0.08\linewidth}{\scriptsize this bird has a that white and and white a , and beak .} &

        \parbox[c]{0.09\linewidth}{\scriptsize the bird has black overall feathers and thick a bright eyes .} &

        \parbox[c]{0.11\linewidth}{\scriptsize this bird has a black head , a and breast and and a a webbed well and orange beak .} 
        \\[2ex]

        \cmidrule(lr){1-4} \cmidrule(lr){5-8}

        \raisebox{13ex}{\parbox[c]{0.09\linewidth}{\small this bird is mostly black with a red and white stripe at the base of his wings .}} &
        \includegraphics[width=0.07\linewidth]{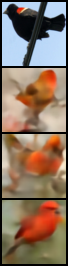} &

        \includegraphics[width=0.07\linewidth]{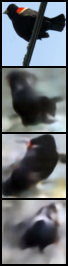} &

        \includegraphics[width=0.07\linewidth]{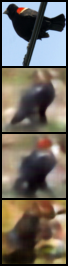} &
        
        \raisebox{14ex}{\parbox[c]{0.10\linewidth}{\small an average sized bird with a black nape and yellow body feathers .}} &
        \includegraphics[width=0.07\linewidth]{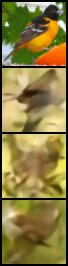} &

        \includegraphics[width=0.07\linewidth]{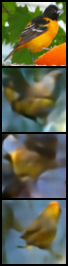} &

        \includegraphics[width=0.07\linewidth]{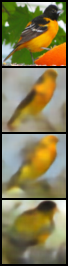} \\
    \end{tabular}
    \vspace*{-2ex}
    \caption{Cross-modality generation on CUB.
    We combine posterior sample $\rvz$ of the modality being conditioned on, with prior sample $\rvw$ of the other modality for generation. 
    \textbf{Top row}: image-to-text. 
    \textbf{Bottom row}: text-to-image. The top image is the original image paired with text, while the rest three are different samples.
    Note our generations better match the conditional input in color.}
    \label{fig:cub_cross_generation}
    \vspace*{-0ex}
\end{figure}

We use ResNet as encoders and decoders for images, while convolution network as those for texts (using one-hot representation of text with a vocabulary of 1,590 words). 
And the dimensionality is set to 48 for $\rvz$ and 16 for $\rvw_m$, following~\citet{palumbo2023mmvaeplus}.
After representation learning, we perform latent linear classification similar to the previous section. With disentangled latent representations, the target (random) classification accuracy is 50\% for predicting direction from $\rvz$, and 12.5\% for predicting category from $\rvw_1$ (derived from image). The results of latent classification are given in Table~\ref{tab:cub_results_part1}. Again, cross-view mutual information maximization is critical for recovering $\rvz$, when we do not have a very strong likelihood model (due to limited image data). On the other hand, generative augmentation still helps reduce redundancy in latent space.
In Figure~\ref{fig:cub_cross_generation}, we provide examples of cross-modality generations and our method achieves more coherent generation than MMVAE+;
additional conditional generations are given in Appendix~\ref{sec:cub_cond_gen}.
{We note that DisentangledSSL performs well for extracting $\rvz$ (their first step has a objective that similarly maximizes mutual information across views), but failed to retain private information in its second step. In contrast, our model keeps the most useful information in the latent space with generative modeling.}

\subsection{Results on The Cancer Genome Atlas (TCGA)}
\vspace*{-1ex}

TCGA dataset\footnote{https://www.cancer.gov/tcga} is a real-world multi-omics dataset that is by nature multimodal. Using the same data processing procedure from~\cite{lee2021variational}, we obtain a dataset of 10,960 samples (of which 9,477 are labeled) with 5 views (each of 100 dimensions), each representing a molecular modality and labels (see Appendix~\ref{sec:dataset_tcga} for details). The binary label represents 1-year mortality of a patient-sample. 
We selected 2 views (mRNA and miRNA) which had 9,874 samples out of all possible combinations after filtering out samples with missing values. After adding data with missing labels to the training set, a 90\%/5\%/5\% split was performed with 5 different seeds. Due to the complex nature of biological data, private information may be predictive as well.

As shown in Table \ref{tab:tcga_results_part2}, our method in general performs better than baseline methods in terms of accuracy (see 
Appendix~\ref{sec:tcga_accuracy} for AUROC results), both learned shared and private latent spaces are predictive, and combining $\rvz$ and $\rvw$ achieves the best performance. This is likely because clean disentanglement separates predictive information between shared and private latent variables, making predictions based on combined latent space more robust. In particular, $\gL_{\text{CrossMI}}$ contributed most to the performance, and adding diffusion priors in latent space consistently improves performance.

\begin{table}[t]
\centering
\caption{Prediction accuracy on TCGA, averaged over 2 modalities and 5 splits.}
\vspace*{-1.5ex}
\label{tab:tcga_results_part2}
\begin{tabular}{l|c|c|c}
\hline
Model & $\rvz\uparrow$ & $\rvw$ 
& $\rvz+\rvw\uparrow$ \\
\hline
MMVAE &0.695$\pm$0.010  & ---&--- \\
MoPoE-VAE  & 0.695$\pm$0.014 &---&---  \\
DMVAE & 0.688$\pm$0.018 & 0.691$\pm$0.014 &0.697$\pm$0.016 \\
MMVAE+ & 0.692$\pm$0.010&0.690$\pm$0.012& 0.690$\pm$0.011  \\
DisentangledSSL & 0.691$\pm$0.011 &0.691$\pm$0.012 &0.690$\pm$0.011 \\
\hline
IDMVAE \textbf{(ours)}& 0.707$\pm$0.016  &0.708$\pm$0.013&0.718$\pm$0.017  \\
\hspace{0.8em} – $\gL_{\text{CrossMI}}$ ($\lambda_1=0$)  & 0.691$\pm$0.014  & 0.689$\pm$0.010 & 0.691$\pm$0.014\\
\hspace{0.8em} – $\gL_{\text{GenAug}}$ ($\lambda_2=0$) &0.701$\pm$0.015  &0.706$\pm$0.019&0.723$\pm$0.013 \\
+ Diffusion prior & \textbf{0.714$\pm$0.009} &\textbf{0.719$\pm$0.024} &\textbf{0.731$\pm$0.019} \\
\hline
\end{tabular}
\end{table}

%% file: 5_conclusion.tex
\section{Conclusions}
\label{sec:conclusion}
\vspace*{-1ex}

We have proposed IDMVAE, a generative model for learning disentangled representation from multimodal data.
Our innovations include the incorporation of cross-view mutual information maximization for shared variable extraction, redundancy removal based on generative augmentation, and flexible latent priors with diffusion models. 
These components are complimentary to each other and jointly overcome the limitations of pure likelihood modeling, resulting in superior performance than existing state-of-the-art multimodal VAEs as well as non-generative disentanglement method.

In the future, we would like to extend the model to handle missing modalities, leveraging the controllable generation capability of our model. On the other hand, for the CUB dataset, we were not able to generate very high fidelity samples of images, perhaps due to limited data volume and capacity of the decoder. We would like to introduce (possibly pre-trained) diffusion models in the input space to produce high quality samples, which may be more useful for generative augmentation.

%% file: 6_appendix.tex

\clearpage

\clearpage

\section{Different Implementations for Generative Augmentation}
\label{sec:gen-aug-comparison}

In section~\ref{sec:generative-augmentation}, we have discussed two implementations of generative augmentation for redundancy removal.
Here we provide a detailed comparison of them.  

Recall that $\rvx_m$ and $\rvx_m^\prime$ are two input samples, $(\rvz_m, \rvw_m)$ is pair of samples drawn from the posteriors $q(\rvz|\rvx_m)$ and $q(\rvw_m|\rvx_m)$ respectively, and similarly $(\rvz_m^\prime, \rvw_m^\prime)$ is a pair of samples drawn from conditional posteriors for $\rvx_m^\prime$.
With disentanglement and a good generative model, we could independently vary one variable while keeping the other the same to obtain a new sample.
In particular, a sample $\rvx_m^+ \sim p(\rvx_m|\rvz_m,\rvw_m^\prime)$ would share the same $\rvz$ with $\rvx_m$. In turn, when we map $\rvx_m^+$ back to the latent space, $q(\rvz|\rvx_m^+)$ and $q(\rvz|\rvx_m)$ should be similar. 
Likewise, $q(\rvw_m|\rvx_m^+)$ and $q(\rvw_m|\rvx_m^\prime)$ should be similar.

\paragraph{Least squares matching.}
In the first implementation, we would like to minimize $I(\rvz_m; \rvx_n)$ by approximately minimizing $H (\rvw_m|\rvx_m^\prime)$:
\begin{align*}
\E_{\rvx_m\sim p(\rvx_m), \rvx_m^\prime\sim p(\rvx_m), \rvz_m\sim q(\rvz|\rvx_m), \rvw_m^\prime\sim p(\rvw_m|\rvx_m^\prime), \rvx_m^+ \sim p(\rvx_m|\rvz_m,\rvw_m^\prime)}[-\log q(\rvw_m^\prime|\rvx_m^+)].
\end{align*}
Assuming that posteriors are parameterized Gaussians, $\gL_{\text{GenAug},\rvw_m}$ reduces to $\ell_2$ loss for matching means of posteriors, and we implement it as 
\begin{align*}
\gL_{\text{GenAug}}^{lsq} = \E_{\rvx_m\sim p(\rvx_m), \rvx_m^\prime\sim p(\rvx_m), \rvz_m\sim q(\rvz|\rvx_m), \rvw_m^\prime\sim q(\rvw_m|\rvx_m^\prime), \rvx_m^+ \sim p(\rvx_m|\rvz_m,\rvw_m^\prime)}
\norm{\overline{\rvw}_m^\prime - \overline{\rvw}_m^{\prime\prime}}^2
\end{align*}
where $\overline{\rvw}_m^\prime$ is the posterior mean of $q(\rvw_m|\rvx_m^\prime)$ while $\overline{\rvw}_m^{\prime\prime}$ is the posterior mean of $q(\rvw_m|\rvx_m^+)$.

\paragraph{Contrastive matching.} In practice, we find it more stable to use a contrastive loss for matching, i.e.,
\begin{align*}
\gL_{\text{GenAug}}^{contrast} := - Contrast(\rvw_m^{\prime\prime}, \rvw_m^\prime)
\qquad\text{where}\quad 
\rvx_m^+ \sim p(\rvx_m|\rvz_m,\rvw_m^\prime),\;
\rvw_m^{\prime\prime} \sim q(\rvw_m | \rvx_m^+)
.
\end{align*}

We plug in the two different implementations into our loss. In Table~\ref{tab:AugMI_losses_comparison_polymnist_results_part1} and ~\ref{tab:AugMI_losses_comparison_polymnist_results_part2}, we provide the comparison of the two on PolyMNIST-Quadrant, each with its loss coefficient tuned on the validation set.
We find the best coefficients to be $\lambda_1$=80 and $\lambda_2^{lsq}$=0.75 for $\gL_{\text{GenAug}}^{lsq}$, and $\lambda_1$=80 and $\lambda_2^{contrast}$=20 for $\gL_{\text{GenAug}}^{contrast}$; diffusion prior loss has a coefficient of $1.0$ when incorporated.
 We observe that both implementations improve the disentanglement compared with using $\gL_{\text{CrossMI}}$ only, with $\gL_{\text{GenAug}}^{contrast}$ outperforming $\gL_{\text{GenAug}}^{lsq}$.

\begin{table}[htbp]
\centering
\caption{Comparison of $\gL_{\text{GenAug}}^{lsq}$ and $\gL_{\text{GenAug}}^{contrast}$ for generative augmentation regularization in latent linear classification on PolyMNIST-Quadrant. 
Accuracies are averaged over 5 modalities.
}
\label{tab:AugMI_losses_comparison_polymnist_results_part1}
\begin{tabular}{l|cc|cc}
\hline
Our Models & $z \to Digit \uparrow$ & $z \to Quad \downarrow$ & $w \to Quad \uparrow$ & $w \to Digit \downarrow$ \\  
\hline
$\gL_{\text{CrossMI}}$ Only ($\lambda_2=0$)     & 0.977 & 0.277 & 0.999 & 0.202     \\  
\hline
$\gL_{\text{CrossMI}}$ + 
    $\gL_{\text{GenAug}}^{lsq}$                 & 0.972 & 0.267 & 0.999 & 0.186     \\  
\hspace{0.8em} + diffusion prior                & 0.980 & \bf 0.263 & 0.999 & 0.154     \\  
\hline
$\gL_{\text{CrossMI}}$ + 
    $\gL_{\text{GenAug}}^{contrast}$            & \bf 0.983 & 0.271 & 0.999 & 0.162     \\  
\hspace{0.8em} + diffusion prior                & 0.982 & 0.267 & 0.999 & \bf 0.143     \\  
\hline
\end{tabular}
\end{table}

\begin{table}[htbp]
\centering
\caption{Comparison of $\gL_{\text{GenAug}}^{lsq}$ and $\gL_{\text{GenAug}}^{contrast}$ for generative augmentation regularization in generative coherence, averaged over 5 views, on PolyMNIST-Quadrant. We use subscript $q$ to indicate samples from posteriors and subscript $p$ to indicate samples from priors.
For generated images, digit label is determined by $\rvz_{s,q}$ or otherwise random (with target accuracy 10\%), quadrant label is determined by $\rvw_{s,q}$ or otherwise random (with target accuracy 25\%).  
}
\label{tab:AugMI_losses_comparison_polymnist_results_part2}
\begin{tabular}{@{}l||@{\hspace{0\linewidth}}c@{\hspace{0\linewidth}}c@{\hspace{0\linewidth}}|@{\hspace{0\linewidth}}c@{\hspace{0\linewidth}}c@{\hspace{0\linewidth}}||@{\hspace{0\linewidth}}c@{\hspace{0\linewidth}}c@{\hspace{0\linewidth}}||@{\hspace{0\linewidth}}c@{}}
\hline
\multirow{3}{*}{Our Models} & \multicolumn{4}{c@{\hspace{0.01\linewidth}}||}{Self Gen ($s=t$)} & \multicolumn{2}{c@{\hspace{0.01\linewidth}}||}{Cross Gen  ($s \neq t$)} & Uncond. \\
\cline{2-8}
& \multicolumn{2}{c|}{$Gen(\rvz_{s,q}, \rvw_{t,p})$} & \multicolumn{2}{c||}{$Gen(\rvz_{t,p}, \rvw_{s,q})$} & \multicolumn{2}{c||}{$Gen(\rvz_{s,q}, \rvw_{t,p})$} & $Gen(\rvz_{p}, \rvw_{p})$\\
\cline{2-8}
 & $Digit \uparrow$ & $ Quad \downarrow$ & $ Digit \downarrow$ & $ Quad \uparrow$ & $ Digit \uparrow$ & $ Quad \downarrow$ & $ Digit \uparrow$ \\
\hline
$\gL_{\text{CrossMI}}$ Only ($\lambda_2=0$)     & 0.670         & 0.250 & 0.370         & 0.999 & 0.671         & 0.250  & 0.008        \\
\hline
$\gL_{\text{CrossMI}}$ + 
    $\gL_{\text{GenAug}}^{lsq}$                 & 0.817         & 0.250 & 0.219         & 0.999 & 0.812         & 0.250  & 0.044    \\ %
\hspace{0.8em} + Diffusion prior                 & 0.917         & 0.249 & 0.109         & 0.999 & 0.875         & 0.250  & \bf 0.668    \\
\hline
$\gL_{\text{CrossMI}}$ + 
    $\gL_{\text{GenAug}}^{contrast}$            & 0.898         & 0.249 & 0.162         & 0.999 & 0.881         & 0.250  & 0.070        \\
\hspace{0.8em} + Diffusion prior                &\textbf{0.942} & 0.251 &\textbf{0.106} & 0.999 &\textbf{0.887} & 0.251  & 0.664    \\
\hline
\end{tabular}
\end{table}

\clearpage

\section{Details and Additional Results on PolyMNIST-Quadrant}

%

\subsection{Conditional Generation}
\label{sec:polymnist_quadrant_conditional_generation}

\begin{figure}[h]
    \centering

    \begin{tabular}{@{}ccc@{}}
        \begin{minipage}[t]{0.31\linewidth}
            \centering
            \includegraphics[width=\linewidth]{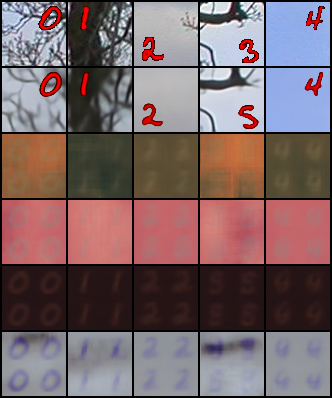}
            \vspace{1mm}
             MMVAE \\
        \end{minipage} &
        \begin{minipage}[t]{0.31\linewidth}
            \centering
            \includegraphics[width=\linewidth]{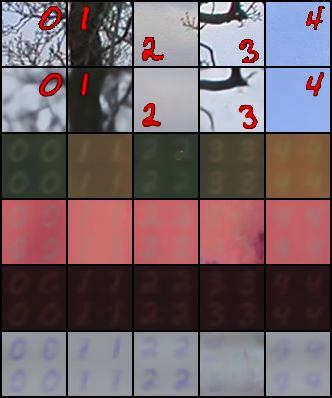}
            \vspace{1mm}
             MoPoE-VAE \\
        \end{minipage} &
        \begin{minipage}[t]{0.31\linewidth}
            \centering
            \includegraphics[width=\linewidth]{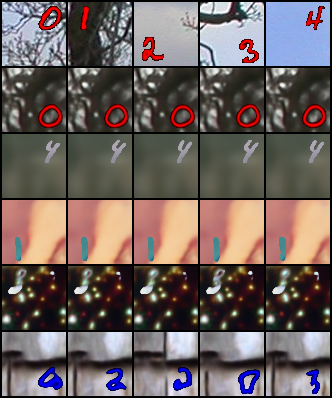}
            \vspace{1mm}
             DMVAE \\
        \end{minipage} \\[5ex]
        \begin{minipage}[t]{0.31\linewidth}
            \centering
            \includegraphics[width=\linewidth]{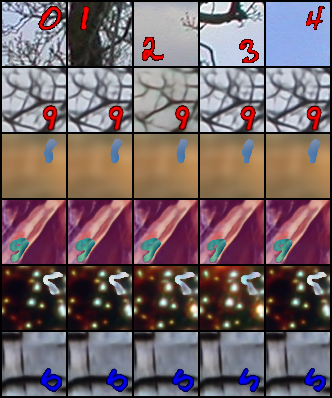}
            \vspace{1mm}
             MMVAE+ \\
        \end{minipage} &
        \begin{minipage}[t]{0.31\linewidth}
            \centering
            \includegraphics[width=\linewidth]{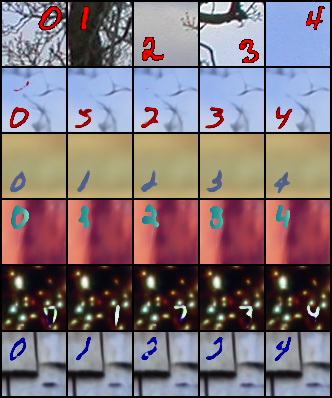}
            \vspace{1mm}
             IDMVAE \\
        \end{minipage} &
        \begin{minipage}[t]{0.31\linewidth}
            \centering
            \includegraphics[width=\linewidth]{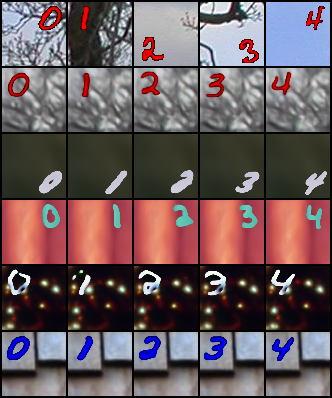}
            \vspace{1mm}
             + diffusion prior
             \\
        \end{minipage}
    \end{tabular}
    \caption{
    Conditional generations on PolyMNIST-Quadrant, with conditioning on $\rvz$. The top row shows the samples (from modality 1) we condition on. We sample $\rvz \sim q(\rvz|\rvx_1)$, sample the private variable from the corresponding prior $\rvw_m \sim p(\rvw_m|\rvx_m)$, and generate a new sample from $p(\rvx_m|\rvz,\rvw_m)$. 
    Row 2 to row 6 are generated samples for modalities 1 to 5. Note for well-disentangled latent variables, each column shall contain the same digit $\rvz$. {For each row we used the same prior sample of $\rvw$, so images in the same row shall have the same quadrant, writing style, and background.}
    }
    \label{fig:conditional_m0_Qz_Pw_comparison}
\end{figure}

\begin{figure}[t]
    \centering
    \begin{tabular}{@{}c@{\hspace{0.02\linewidth}}c@{\hspace{0.02\linewidth}}c@{\hspace{0.02\linewidth}}c@{}}
        \begin{minipage}[t]{0.23\linewidth}
            \centering
            \includegraphics[width=\linewidth]{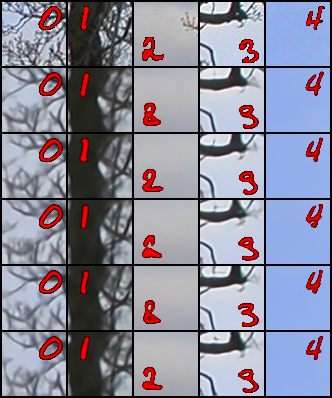}
            \vspace{1mm}
             DMVAE 
             %
        \end{minipage} &
        \begin{minipage}[t]{0.23\linewidth}
            \centering
            \includegraphics[width=\linewidth]{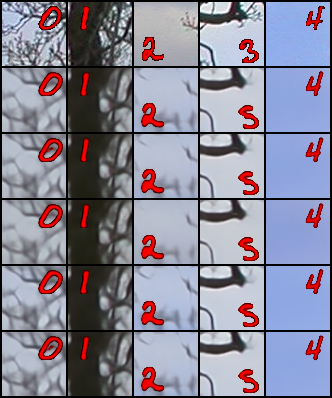}
            \vspace{1mm}
             MMVAE+ 
             %
        \end{minipage} &
        \begin{minipage}[t]{0.23\linewidth}
            \centering
            \includegraphics[width=\linewidth]{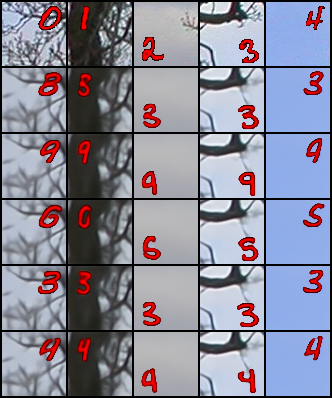}
            \vspace{1mm}
             IDMVAE 
             %
        \end{minipage} &
        \begin{minipage}[t]{0.23\linewidth}
            \centering
            \includegraphics[width=\linewidth]{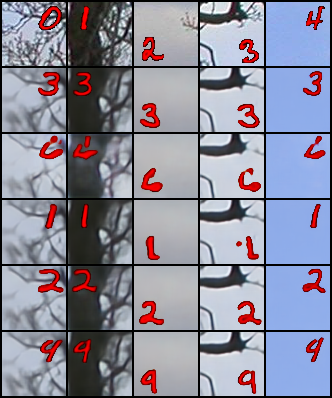}
            \vspace{1mm}
              + diffusion prior
             %
        \end{minipage}
    \end{tabular}
    \caption{Conditional generations on PolyMNIST-Quadrant, with conditioning on $\rvw_m$. The top row shows the samples (from modality 1) we condition on.
    We sample $\rvw \sim q(\rvw|\rvx_1)$, and sample $\rvz \sim p(\rvz)$, and generate a new sample from $p(\rvx_1|\rvz,\rvw_1)$.
    Row 2 to row 6 are generated samples.
    Note for well-disentangled latent variables, each column shall have the same quadrant position and background. {For each row we used the same prior sample of $\rvz$, so images in the same row shall have the same digit.}
    }
    \label{fig:conditional_m0_Qw_Pz_comparison}
\end{figure}

\begin{figure}[t]
    \centering
    \begin{tabular}{@{}c@{\hspace{0.05\linewidth}}c@{}}
        \includegraphics[width=0.45\linewidth]{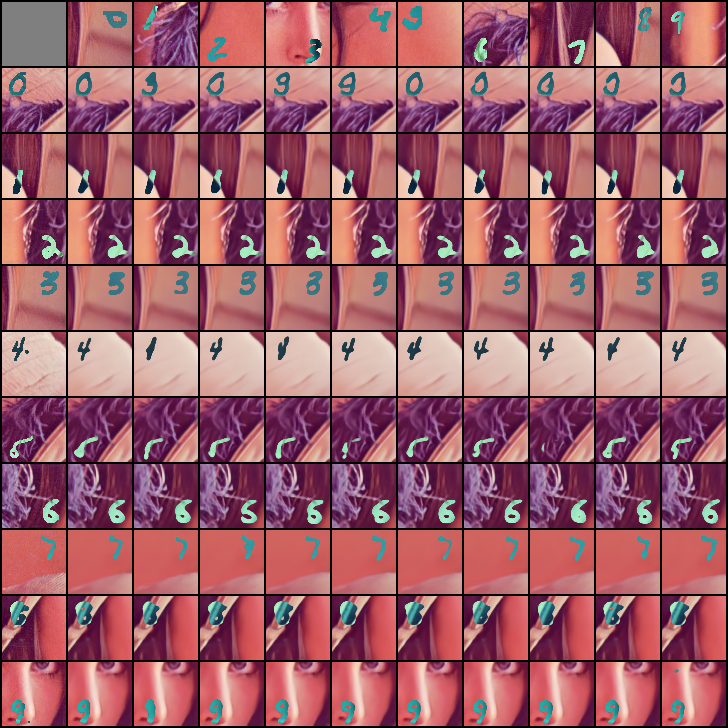} &
        \includegraphics[width=0.45\linewidth]{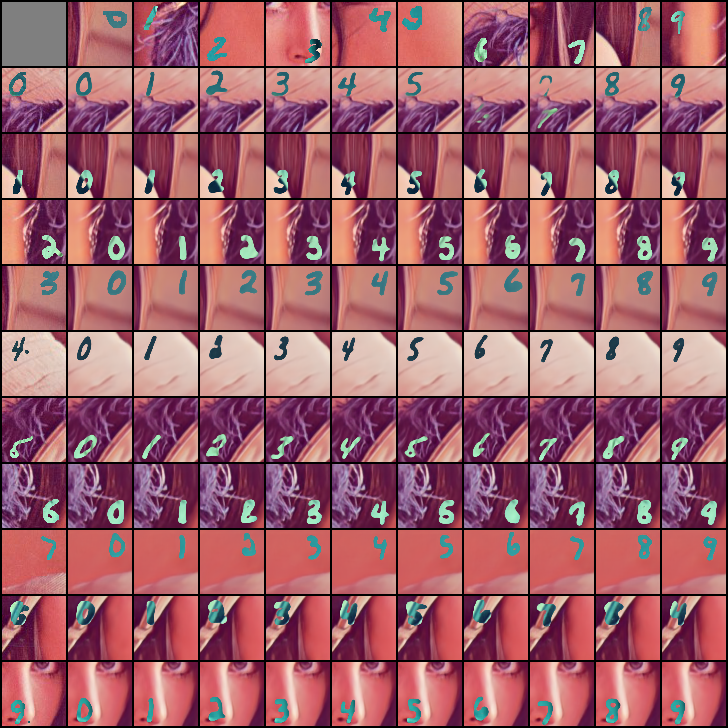} \\
        MMVAE+ & IDMVAE  \\
    \end{tabular}
    \caption{Conditional generation on PolyMNIST-Quadrant. The first row and first column contain images for which we extract posterior samples of $\rvz$ and $\rvw_m$, respectively. And the rest of the grid contains generated images using latent samples of the corresponding row and column.
    This figure illustrates the samples we use in generative augmentation.
    }
    \label{fig:conditional_m2_Qz_Qw_comparison}
\end{figure}

In Figure~\ref{fig:conditional_m0_Qz_Pw_comparison} and Figure~\ref{fig:conditional_m0_Qw_Pz_comparison}, we provide additional results on conditional generations, where one latent variable is sampled from the posterior, while the other is sampled from the prior.
In Figure~\ref{fig:conditional_m2_Qz_Qw_comparison}, we provide conditional generations for which both $\rvz$ and $\rvw$ are sampled from posteriors; this simulates the samples we use in $\gL_{\text{GenAug}}$.
In all cases, our method provides the most coherent generations, consistent with the quantitative results in Section~\ref{sec:polymnist_results}.


%


%


\subsection{Implementation Details}
\label{sec:polymnist_quadrant_implementation_details}
We utilize a deep residual network (ResNet) architecture of $3$ residual blocks, with the number of filters doubling from 64 to up to 512 after each block for the encoder, 
and the number of filters halving after each decoder for the decoder, for all five modalities. And each modality's information is factorized in the latent space into a shared latent dimension of 32 and a private latent dimension of 128. Models are trained for 100 epochs using the Adam optimizer with a learning rate of $5e^{-4}$ and a batch size of 128, and use the other default hyperparameters of MMVAE+ baseline, including the KL divergence coefficient $\beta$ of 2.5. 
We performed a grid search over the coefficients to tune the regularization terms, $\lambda_1$ and $\lambda_2$, after training for 100 epochs. 
We search them in the range $[0.01, 100]$.
We tune $\lambda_1$ individually first to find the best general performance in latent classification for $\gL_{\text{CrossMI}}$, and fix the $\lambda_1$, then combine with $\gL_{\text{GenAug}}^{contrast}$, and find the best combination of $\lambda_1$=80 and $\lambda_2^{contrast}$=20. Finally, we tune the diffusion prior weight to 1.0 out of \{0.01, 0.1, 1.0, 10.0\}, which optimizes the final general performance at the 100th epoch. 

\clearpage

\subsection{Latent Visualization}
\label{sec:polymnist-quadrant-latent-visualization}

\begin{figure}[h!]
    \centering
    \begin{tabular}{@{}cc@{}}
        \begin{minipage}[t]{0.45\linewidth}
            \centering
            \includegraphics[width=\linewidth]{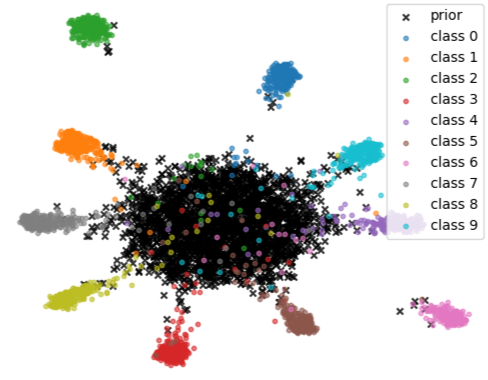}
            \vspace{1mm}
             w.o. diffusion prior
        \end{minipage} &

        \begin{minipage}[t]{0.45\linewidth}
            \centering
            \includegraphics[width=\linewidth]{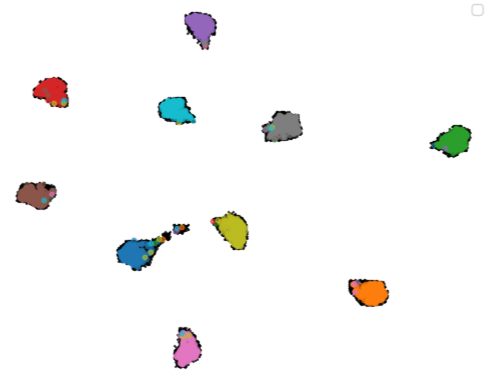}
            \vspace{1mm}
             w. diffusion prior
        \end{minipage}
    \end{tabular}    
    \caption{
    2D Latent Visualization (UMAP) of learned shared latent representation $\rvz$ by our methods on the test set. We color each representation according to its ground truth digit label. Black points are samples from the prior. 
    }
    \label{fig:conditional_m0_Qz_Pz_2D}
\end{figure}

In Figure~\ref{fig:conditional_m0_Qz_Pz_2D}, we provide 2D visualizations of the latent representation $q(\rvz|\rvx_1)$ on the PolyMNIST-Quadrant test set.
Observe that, without diffusion prior (left panel), there exists a gap between the posterior and the Gaussian prior, whereas the capacity of diffusion prior is strong enough to ensure good overlap between the two distributions (right panel). 

\clearpage

\section{Details and Additional Results on CUB}

\label{sec:dataset_cubcluster}

\subsection{Dataset}
\begin{figure}[htbp]
    \centering
    \begin{tabular}{@{}c@{\hspace{0.0\linewidth}}c@{}}
        \includegraphics[width=1.0\linewidth]{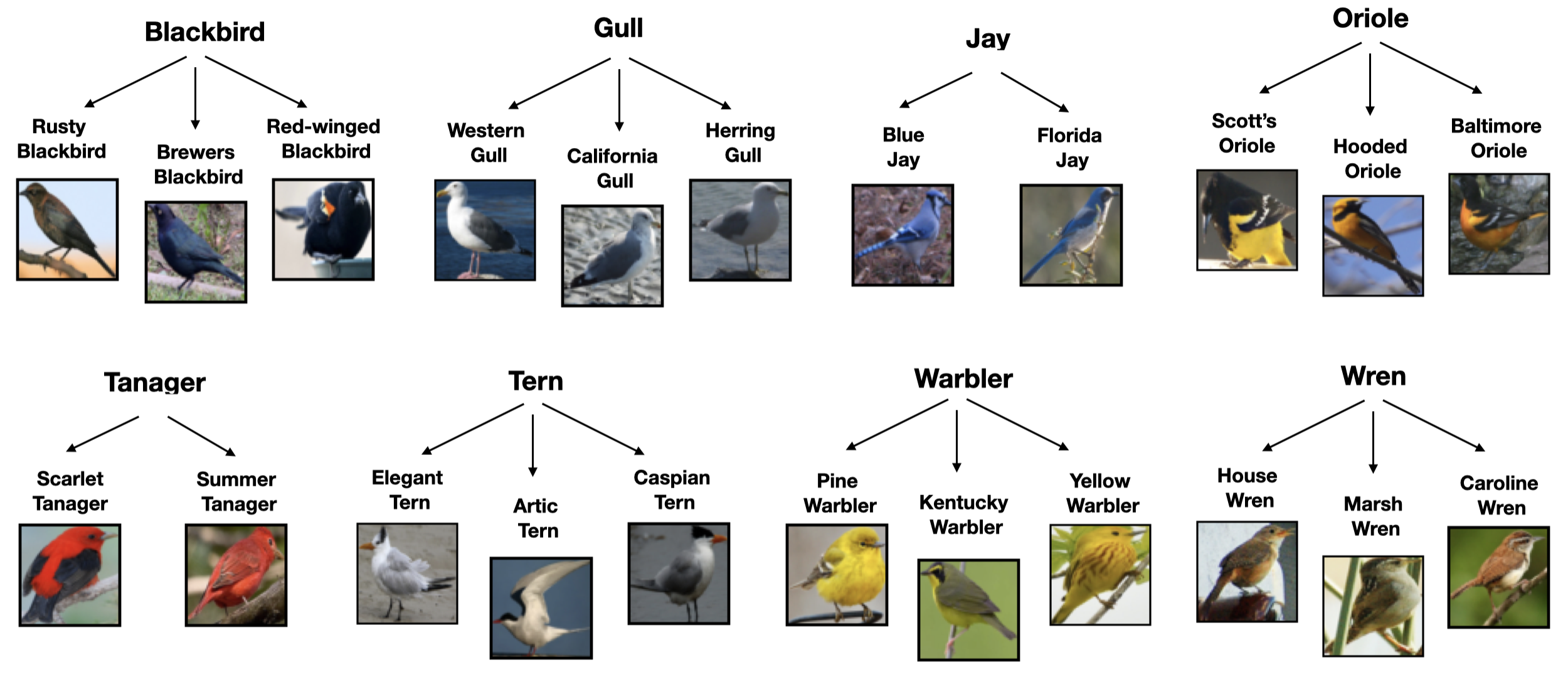} \\
    \end{tabular}
    \caption{CUB category labels: dividing 22 species into 8 super-categories~\citep{palumbo2024deep}.
    }
    \label{fig:CUB_dataset_structure}
\end{figure}

The 8 super-categories, namely Blackbird, Gull, Jay, Oriole, Tanager, Tern, Warbler, and Wren, are created following the same grouping method introduced by ~\citet{palumbo2024deep}, as shown in Figure~\ref{fig:CUB_dataset_structure}.

\begin{figure}[htbp]
    \centering
    \begin{tabular}{@{}c@{\hspace{0.01\linewidth}}c@{\hspace{0.01\linewidth}}c@{\hspace{0.0\linewidth}}c@{}}
        \raisebox{-13ex}{
        \begin{minipage}[t]{0.65\linewidth}
            \centering
            \includegraphics[width=\linewidth]{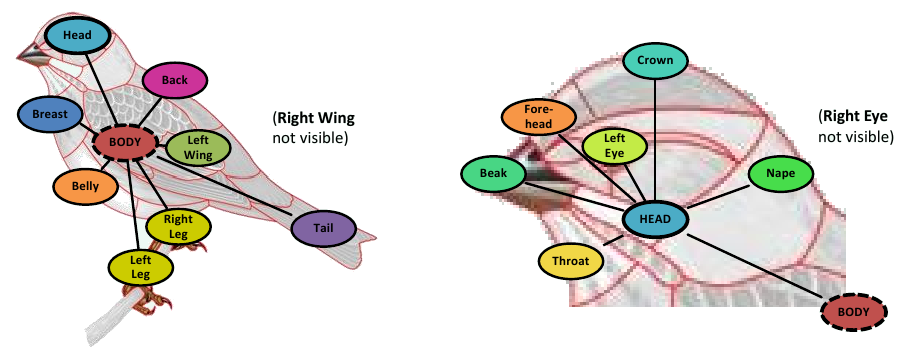}
            \vspace{1mm}
             a. Collected Parts 
        \end{minipage}} &
        \begin{minipage}[t]{0.15\linewidth}
            \centering
            \includegraphics[width=\linewidth]{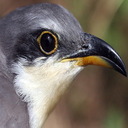}
            \includegraphics[width=\linewidth]{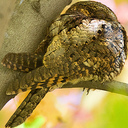}            
            \vspace{1mm}
             b. Invisible
        \end{minipage} &
        \begin{minipage}[t]{0.15\linewidth}
            \centering
            \includegraphics[width=\linewidth]{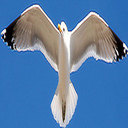}
            \includegraphics[width=\linewidth]{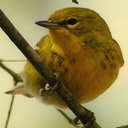}
            \vspace{1mm}
             c. Ambiguous
        \end{minipage}
    \end{tabular}
    \caption{The collected parts in the original dataset~\citep{wah2011caltech}, and sample images with invisible `body' or `head' location annotations or ambiguous horizontal direction. 
    }
    \label{fig:CUB_parts_and_amb_img}
\end{figure}

In addition, we introduce a private binary label representing the bird's horizontal direction, determined by the part location annotations provided in the original dataset~\citep{wah2011caltech}, as illustrated in Figure~\ref{fig:CUB_parts_and_amb_img}.
Specifically, we compare the average horizontal position of the group of the bird's `head' parts with the average horizontal position of the group of its `body' parts. If the head is positioned to the left of the body, the direction label is `left' (label 0); otherwise, it is `right' (label 1). This creates a modality-specific (private) label for the image that cannot be inferred from the text captions. At the same time, as shown in Figure~\ref{fig:CUB_parts_and_amb_img} (b) and (c), %
a very small fraction of the images have 
invisible `head' or `body' location annotations, or the locations are too close, in which case they are not assigned the direction label. 
Direction labels of validation and test images are verified by human.

\subsection{Implementation Details}
For the image modality, we utilize a deep residual network (ResNet) architecture of $5$ residual blocks, with the number of filters doubling from 64 to up to 1024 after each block for the encoder, and the number of filters halving after each block for the decoder.
For text modality, we utilize a convolutional neural network, CNN-based encoder and decoder with one-hot encoded captions from a vocabulary size of 1590 words.
The shared variable has a dimensionality of 48, and the private variable has a dimensionality of 16.
Models are trained for 150 epochs using the Adam optimizer with a learning rate of $10^{-3}$ and a batch size of 128, and other default hyperparameters of the MMVAE+ baseline, including the KL divergence coefficient $\beta=1.0$.

During training, we apply horizontal flip augmentation to the image modality, with a flip probability of $0.5$.
Then we tune the $\gL_{\text{CrossMI}}$, $\gL_{\text{GenAug}}^{contrast}$, and diffusion prior similarly to~\ref{sec:polymnist_quadrant_implementation_details}, and obtain the optimal coefficients $\lambda_1=40$, $\lambda_2^{contrast}=0.05$, and diffusion loss weight $0.1$.

\subsection{Conditional Generation}
\label{sec:cub_cond_gen}

\begin{figure}[h]
    \centering
    \begin{tabular}{ccc}
        MMVAE+ & IDMVAE (Ours) & + diffusion prior \\
        \includegraphics[width=0.3\linewidth]{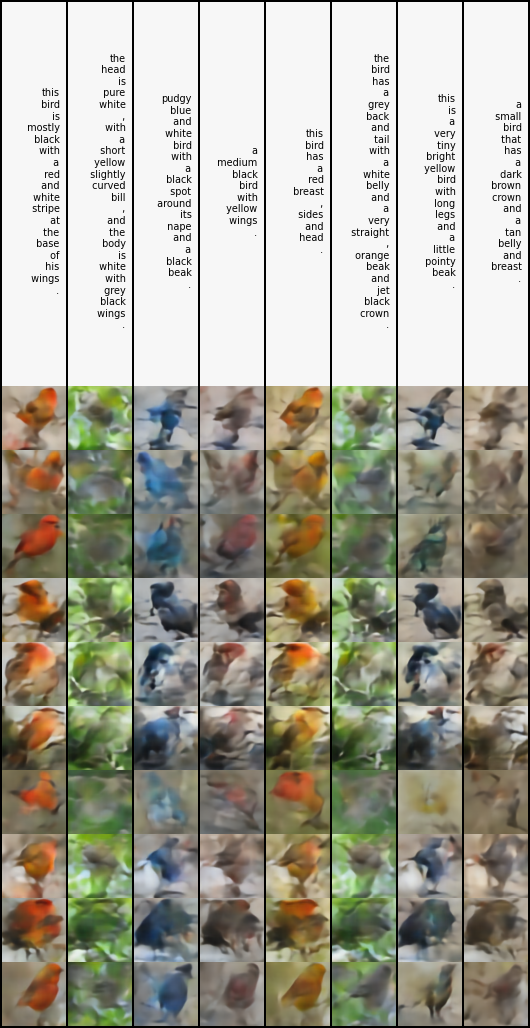} &
        \includegraphics[width=0.3\linewidth]{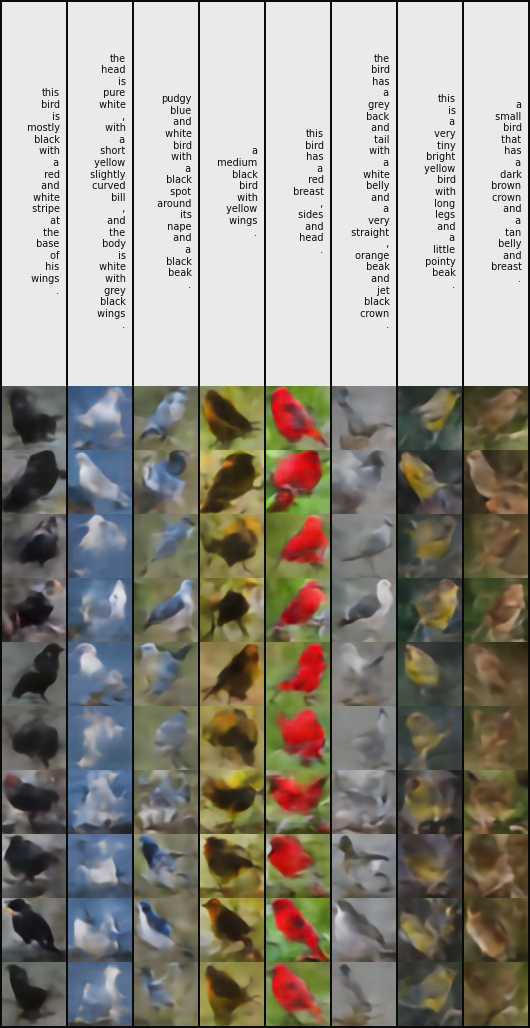} &
        \includegraphics[width=0.3\linewidth]{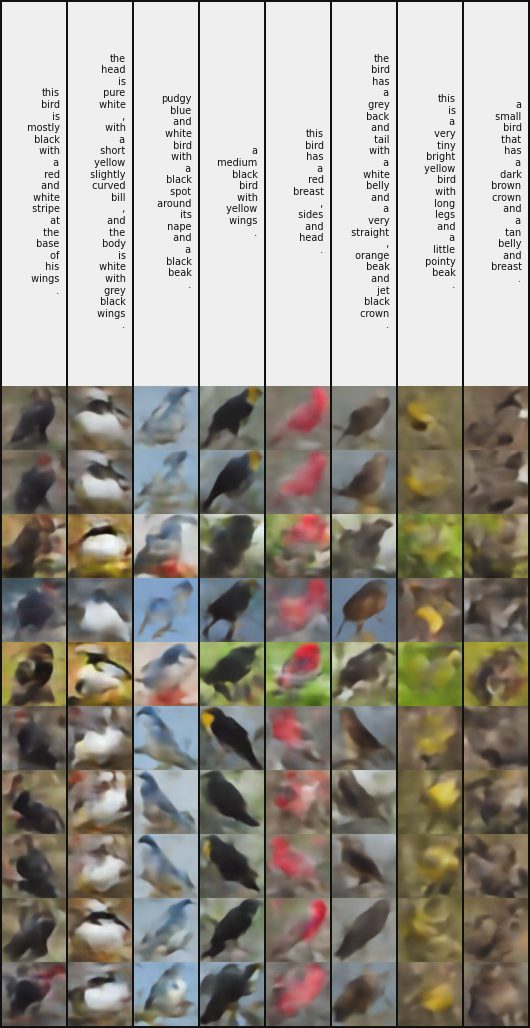} \\
    \end{tabular}
    \caption{
    Text-to-image generation on CUB.
    We combine posterior sample $\rvz$ of the text modality (top row), with prior sample $\rvw$  of the image modality (shared by each row) for generation. 
    }
    \label{fig:cub_image_table_c2i_comparison}
\end{figure}

\begin{figure}[htbp]
    \centering
    \begin{tabular}{ccc}
        MMVAE+ & IDMVAE (Ours) & + diffusion prior \\
        \includegraphics[width=0.3\linewidth]{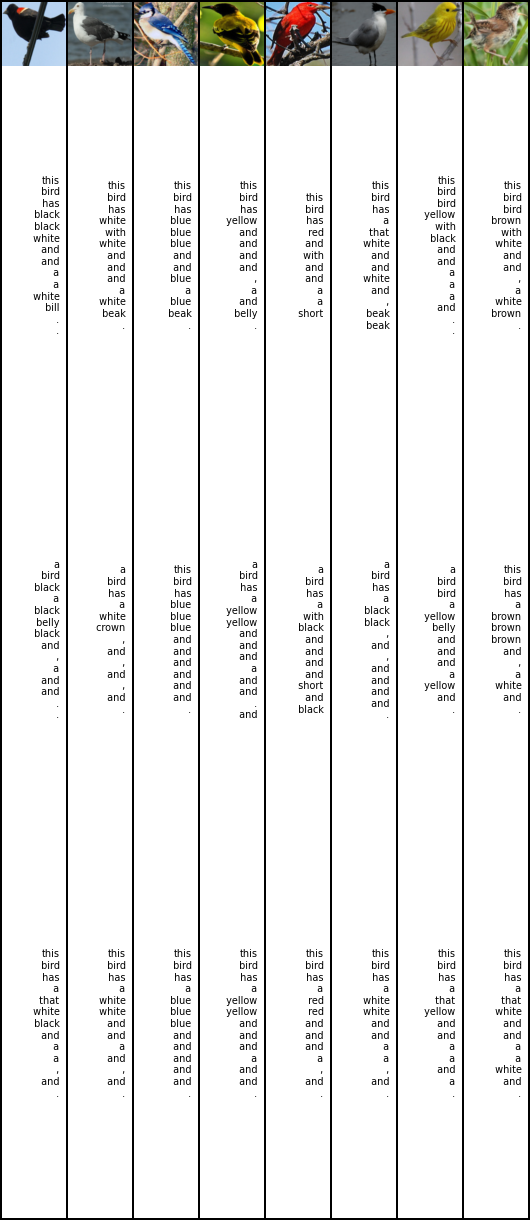} &
        \includegraphics[width=0.3\linewidth]{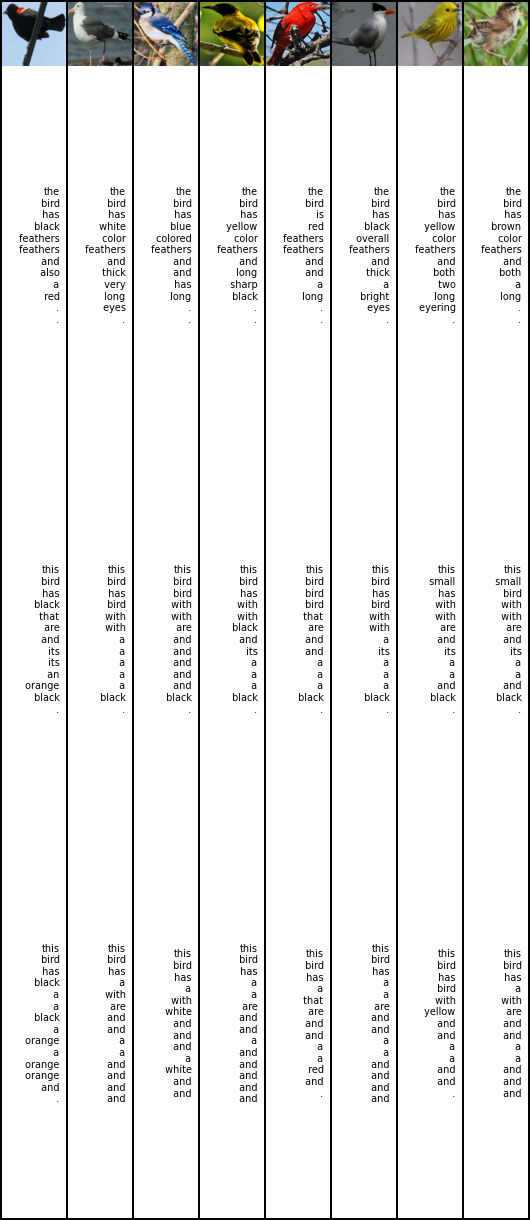} &
        \includegraphics[width=0.3\linewidth]{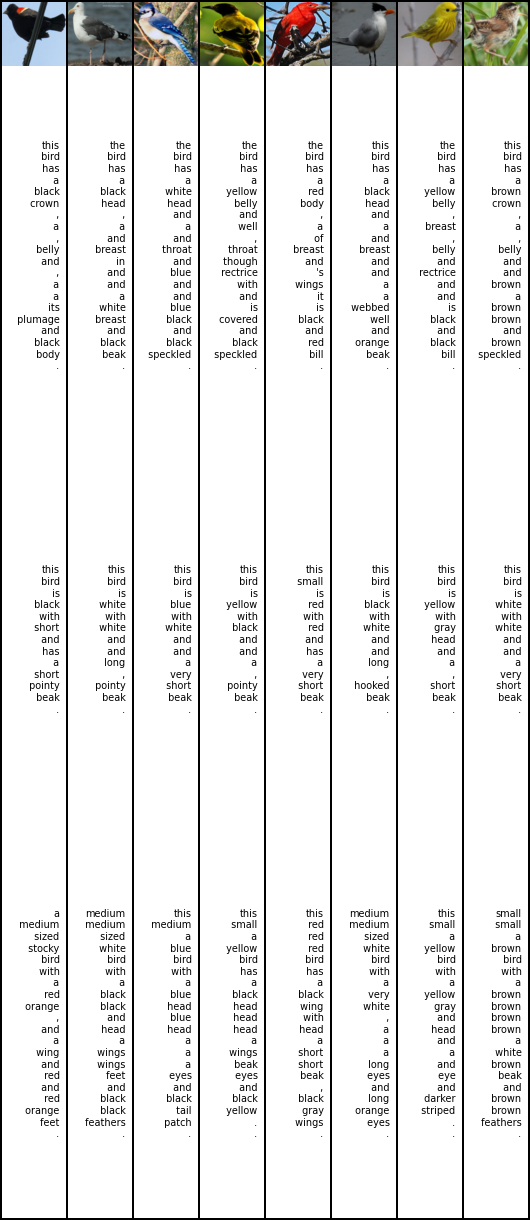} \\
    \end{tabular}
    \caption{
    Image-to-text generation on CUB. We combine posterior sample $\rvz$ of the image modality (top row), with prior sample $\rvw$ of the text modality (shared by each row) for generation.}
    \label{fig:cub_image_table_i2c_comparison}
\end{figure}

\begin{figure}[htbp]
    \centering
    \begin{tabular}{ccc}
        MMVAE+ & IDMVAE (Ours) & + diffusion prior \\
        \includegraphics[width=0.3\linewidth]{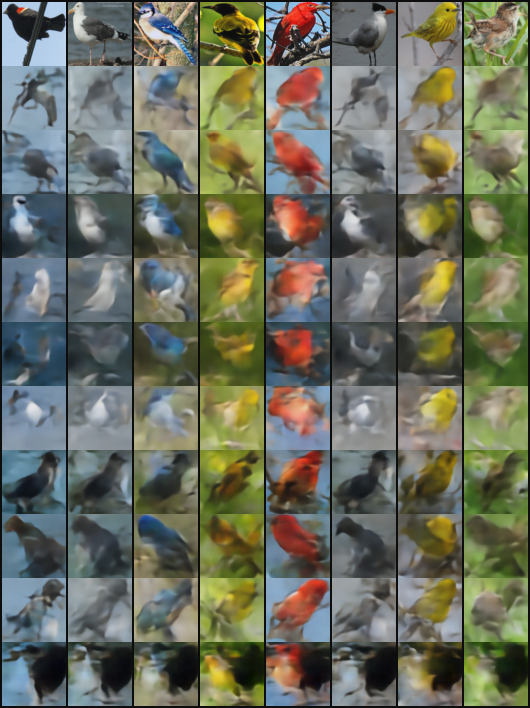} &
        \includegraphics[width=0.3\linewidth]{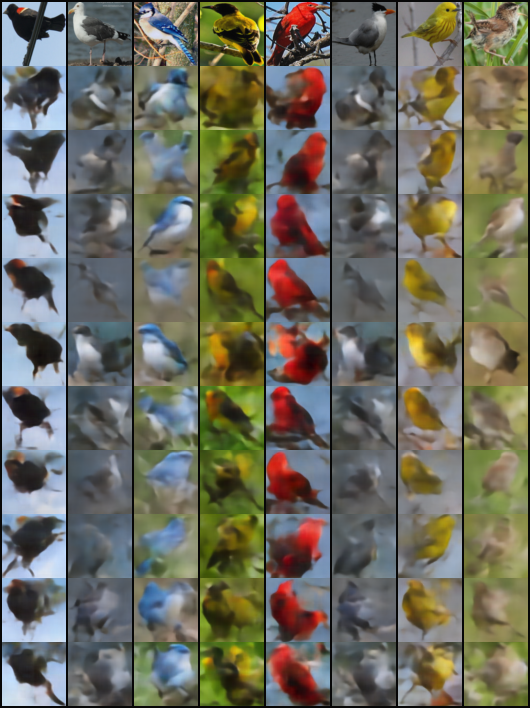} &
        \includegraphics[width=0.3\linewidth]{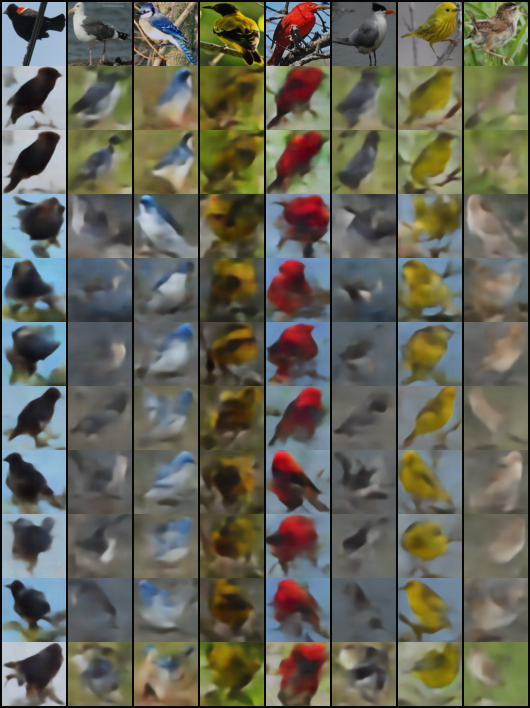} \\
    \end{tabular}
    \caption{Image-to-image generation on CUB.
    We combine the posterior sample $\rvz$ of the image modality (top row), with the prior sample $\rvw$  of the image modality (shared by each row) for generation.
    }
    \label{fig:cub_image_table_i2i_comparison}
\end{figure}

In Figure~\ref{fig:cub_image_table_c2i_comparison}, ~\ref{fig:cub_image_table_i2c_comparison},~\ref{fig:cub_image_table_i2i_comparison}, we provide conditional generation of competitive methods.

\clearpage
\section{Details and Additional results on TCGA}

\subsection{Data Preparation}
\label{sec:dataset_tcga}

TCGA dataset (2016 version) is processed by combining multi-omics data across different cancer types and different patients before features are selected and kernel PCA was performed to reduce the dimensionality to 100. Because clinical trials usually have small sample sizes, high dimensionalities, and complex dependencies, this dataset is ideal to test the robustness of our method. One significant outcome of cancer multi-omics data is the days of survival after samples were collected. Because patients may not show up for checkups, it is possible that a sample is censored (unlabeled). The days of survival are then converted to a binary 1-year mortality indicator. The dataset itself has missing views but contains predictive information and high correlations among different views, making it suitable for our task. Further notice that this dataset does not contain private ground truth.

\subsection{Implementation}
\label{sec:tune_tcga}
A 2-layer MLP with 128 as hidden dimensions was used for encoding and decoding with 48 latent dimensions (16 for $\rvz$ and 32 for $\rvw$). Evaluation for all methods regarding this dataset is done by averaging logits from each view. 50 epochs were run to train the model. For TCGA dataset, baseline methods are performed with default hyperparameter, which gives KL divergence a coefficient of 2.5. For DisentangledSSL baseline in particular, step 1 coefficient was set to 0 since we use posterior mean instead of zsample to match other methods and step 2 coefficient was set to 0.01. To tune our method, we performed a grid search with coefficients $\{0.001, 0.01, 0.1, 1, 10, 100\}$ and chose the best combination on validation set, before recording the performance on test set. $\lambda_1=10,\lambda_2=0.001$ were chosen to be the best combination at 40 epochs. For ablation studies, we set one coefficient to be 0 while keeping the other one optimal in a combined setting. For the optimal coefficients combination, we used the model at epoch 40; $\lambda_1=0$, at epoch 50; and $\lambda_2=0$, at epoch 35. For adding a diffusion prior, we tuned the diffusion weight to be 0.1 out of 0.1, 1, 10 while keeping $\lambda_1,\lambda_2$ same as the optimal combination and chose the best performance at validation set at epoch 40.

\subsection{Prediction AUROC}
\label{sec:tcga_accuracy}

In Table~\ref{tab:tcga_results_part1}, we provide the linear classification AUROC of different methods using latent representations. The relative merits of different methods are consistent with that observed with the accuracy metric in Table~\ref{tab:tcga_results_part2}.

\begin{table}[h!]
\centering
\caption{Prediction AUROC Performance with ablation on TCGA dataset, averaged over 2 modalities and 5 splits. Tuning reported in Appendix \ref{sec:tune_tcga}.
}
\label{tab:tcga_results_part1}
\begin{tabular}{l|c|c|c}
\hline
Model & $\rvz\uparrow$ & $\rvw$ 
& $\rvz+\rvw\uparrow$ \\
\hline
MMVAE &0.653$\pm$0.033   & --- &--- \\ %
MoPoE-VAE & 0.660$\pm$0.024 &--- &---  \\
DMVAE & 0.609$\pm$0.030  & 0.636$\pm$0.037 & 0.643$\pm$0.032 \\
MMVAE+ & 0.586$\pm$ 0.027  &0.581$\pm$0.033  & 0.585$\pm$0.033  \\
DisentangledSSL &0.693$\pm$0.046 &0.551$\pm$0.019 &0.699$\pm$0.045 \\
\hline
IDMVAE \textbf{(ours)} & 0.740$\pm$0.025   &0.740$\pm$0.022 & 0.767$\pm$0.026 \\
\hspace{0.8em} – $\gL_{\text{CrossMI}}$ ($\lambda_1=0$) & 0.549$\pm$0.017 & 0.545$\pm$0.026  & 0.548$\pm$0.026 \\
\hspace{0.8em} – $\gL_{\text{GenAug}}$ ($\lambda_2=0$) &0.740$\pm$0.019  &0.746$\pm$0.022 &0.771$\pm$0.021 \\
+ Diffusion prior & \textbf{0.745$\pm$0.024} & \textbf{0.751$\pm$0.029}  &\textbf{0.772$\pm$0.022} \\
\hline
\end{tabular}
\end{table}